\documentclass{article}

 \usepackage[main, final]{neurips_2026}

\usepackage[utf8]{inputenc} 
\usepackage[T1]{fontenc}    
\usepackage{hyperref}       
\usepackage{url}            
\usepackage{booktabs}       
\usepackage{amsfonts}       
\usepackage{nicefrac}       
\usepackage{microtype}      
\usepackage{multirow,enumitem}
\usepackage{array}
\usepackage{graphicx}
\usepackage[table]{xcolor}
\usepackage{pifont}
\usepackage{float}
\usepackage{tikz}
\usepackage{wrapfig}
\usepackage{placeins}
\usepackage{needspace}
\usepackage{caption}

\makeatletter
\providecommand{\onedot}{\futurelet\@let@token\@onedot}
\providecommand{\@onedot}{\ifx\@let@token.\else.\null\fi\xspace}
\makeatother
\usepackage{colortbl,xspace}
\usepackage[most]{tcolorbox}
\definecolor{recblue}{RGB}{70,130,180}
\definecolor{lightpink}{RGB}{250,241,205}
\definecolor{lightblue}{RGB}{223,234,222}
\definecolor{lightlightgray}{RGB}{220,220,220}
\definecolor{lightgreen}{RGB}{80,180,80}
\definecolor{panelgray}{gray}{0.9}

\newcommand{\cmark}{\textcolor{lightgreen}{\ding{51}}}
\newcommand{\xmark}{\textcolor{red!75}{\ding{55}}}
\newcommand{\gain}[1]{\textcolor{lightgreen}{\scriptsize $\mathbf{\uparrow #1}$}}
\newcommand{\down}[1]{\textcolor{red!75}{ \scriptsize $\mathbf{\downarrow #1}$}}
\newcommand{\smallgain}[1]{\textcolor{lightgreen}{\tiny(+#1)}}
\newcommand{\smalldown}[1]{\textcolor{red!75}{\tiny(-#1)}}
\newcommand{\inlineColorbox}[2]{%
  \begingroup
  \setlength{\fboxsep}{1.5pt}%
  \colorbox{#1}{#2}%
  \endgroup
}

\newcommand{\rebuttal}[1]{\textcolor{black}{#1}}

\newcommand{\visualspace}{\mathcal{V}}
\newcommand{\textspace}{\mathcal{L}}

\newcommand{\image}{I}
\newcommand{\desc}{T}

\newcommand{\imagefeat}{\mathbf{v}}
\newcommand{\textfeat}{\mathbf{u}}

\newcommand{\visualencoder}{f_{\mathtt{img}}}
\newcommand{\textencoder}{f_{\mathtt{txt}}}

\newcommand{\saliency}{\mathcal{S}}
\newcommand{\saliencymap}{A}
\newcommand{\saliencymask}{M}

\newcommand{\maskop}{\mathcal{M}}
\newcommand{\contextop}{\mathcal{C}}

\newcommand{\images}{I^{s}}
\newcommand{\imagec}{I^{c}}

\newcommand{\texts}{T^{s}}
\newcommand{\textc}{T^{c}}

\newcommand{\imagesfeat}{\mathbf{v}^{s}}
\newcommand{\imagecfeat}{\mathbf{v}^{c}}

\newcommand{\textsfeat}{\mathbf{u}^{s}}
\newcommand{\textcfeat}{\mathbf{u}^{c}}

\newcommand{\partitioner}{\Phi}

\newcommand{\temperature}{\tau}

\newcommand{\originalloss}{\mathcal{L}_{\mathrm{ori}}}
\newcommand{\salignloss}{\mathcal{L}_{\mathrm{sal}}}
\newcommand{\contextloss}{\mathcal{L}_{\mathrm{ctx}}}

\newcommand{\anchorloss}{\mathcal{L}_{\mathrm{anc}}}

\newcommand{\totalloss}{\mathcal{L}_{\mathrm{total}}}
\newcommand{\ourloss}{\mathcal{L}_{\mathrm{RoSA}}}
\newcommand{\lambdasal}{\lambda_{\mathrm{sal}}}
\newcommand{\lambdactx}{\lambda_{\mathrm{ctx}}}

\newcommand{\lambdaanc}{\lambda_{\mathrm{anc}}}

\newcommand{\prompt}{\mathtt{P}}
\definecolor{mildblue}{RGB}{36,113,163}
\definecolor{midorange}{RGB}{192,111,34}
\definecolor{severered}{RGB}{176,54,54}
\colorlet{salientzero}{lightpink!35!white}
\colorlet{salienthalf}{lightpink!70!white}
\colorlet{salientfull}{lightpink}
\colorlet{salientzero}{lightpink!35!white}
\colorlet{salienthalf}{lightpink!65!white}
\colorlet{salientonly}{lightpink!85!white}
\colorlet{salientfull}{lightpink}
 
\def\ie{\emph{i.e}\onedot}

\definecolor{salientred}{HTML}{C44E52}
\definecolor{randompurple}{HTML}{8172B3}

\DeclareRobustCommand{\salientlegend}{%
  \tikz[baseline=-0.6ex]{
    \draw[salientred, line width=1.3pt] (0,0) -- (0.9,0);
    \fill[salientred] (0.45,0) circle (2pt);
  }%
}

\DeclareRobustCommand{\randomlegend}{%
  \tikz[baseline=-0.6ex]{
    \draw[randompurple, dashed, line width=1.3pt] (0,0) -- (0.9,0);
    \fill[randompurple] (0.45,0) -- +(0,0.07) -- +(-0.07,-0.05) -- +(0.07,-0.05) -- cycle;
  }%
}

\newcommand{\supp}{\textit{Supp.}\xspace}
\newcommand{\ourmethod}{\textsc{RoSA}\xspace}
\newcommand{\ourmethodfull}{Robust Saliency-aware Alignment\xspace}

\title{Mitigating Saliency Collapse: Robust Saliency-Aware Long-Text Image-Text Alignment}

\author{%
    \textbf{Qiuyu Kong}\textsuperscript{1}
    \quad
    \textbf{Zanxi Ruan}\textsuperscript{2,}\thanks{Corresponding author.}
    \quad
    \textbf{Marco Cristani}\textsuperscript{2,4}
    \quad
    \textbf{Yiming Wang}\textsuperscript{3}
    \\[3pt]
    {\normalfont\small
    \textsuperscript{1}Sapienza University of Rome}
    \quad
    {\normalfont\small \textsuperscript{2}University of Verona}
    \\[1pt]
    {\normalfont\small
    \textsuperscript{3} Fondazione Bruno Kessler}
    \quad
    {\normalfont\small
    \textsuperscript{4}Reykjavik University}
    \\[3pt]
    {\normalfont\footnotesize
    \texttt{qiuyu.kong@uniroma1.it}
    \quad \texttt{zanxi.ruan@univr.it}}
    \\
    {\normalfont\footnotesize
    \texttt{marco.cristani@univr.it}
    \quad \texttt{ywang@fbk.eu}}
    \\[2pt]
}

\begin{document}

\maketitle

\begin{abstract}
Long-text image–text alignment is essential for fine-grained scene understanding, yet existing methods mainly focus on extending context length or improving local alignment. We identify an overlooked failure mode, \textit{saliency collapse}, where models over-rely on text that features visual prominence while underutilizing non-salient context. Consequently, retrieval performance degrades sharply when salient content is reduced, despite that the non-salient context remains informative.
To address this, we propose \ourmethod, a saliency-aware finetuning framework that decomposes image-text pairs into complementary salient and contextual views, enabling joint learning of prominent and contextual semantics. We propose a modality-independent decomposition strategy to combine object-level visual saliency estimation with LLM-based text decomposition to bridge vision and language. 
We employ auxiliary objectives to explicitly align these views, ensuring robust representation with both salient and contextual features.
\ourmethod greatly improves long-text retrieval and demonstrates strong robustness to saliency collapse across benchmarks. Moreover, our auxiliary objectives are plug-and-play, bringing consistent gains to existing tuning methods. Code and checkpoints are available at: https://github.com/intelligolabs/RoSA.

\end{abstract}

\section{Introduction}
\label{sec:introduction}
Long-text image-text alignment establishes semantic correspondence between images and detailed textual descriptions~\citep{docci2024,zhang2024long}, a capability fundamental to fine-grained and complex scene understanding. Since the standard CLIP text encoder is restricted to a 77-token context window~\citep{radford2021learning}, contemporary methods have primarily addressed this bottleneck by extending context lengths~\citep{zhang2024long}, enhancing local spatial alignment~\citep{goal2025}, or incorporating structural priors via auxiliary objectives~\citep{ruan2026StructXLIP}.

\begin{figure}[h]
    \centering
    \includegraphics[width=0.9\linewidth]{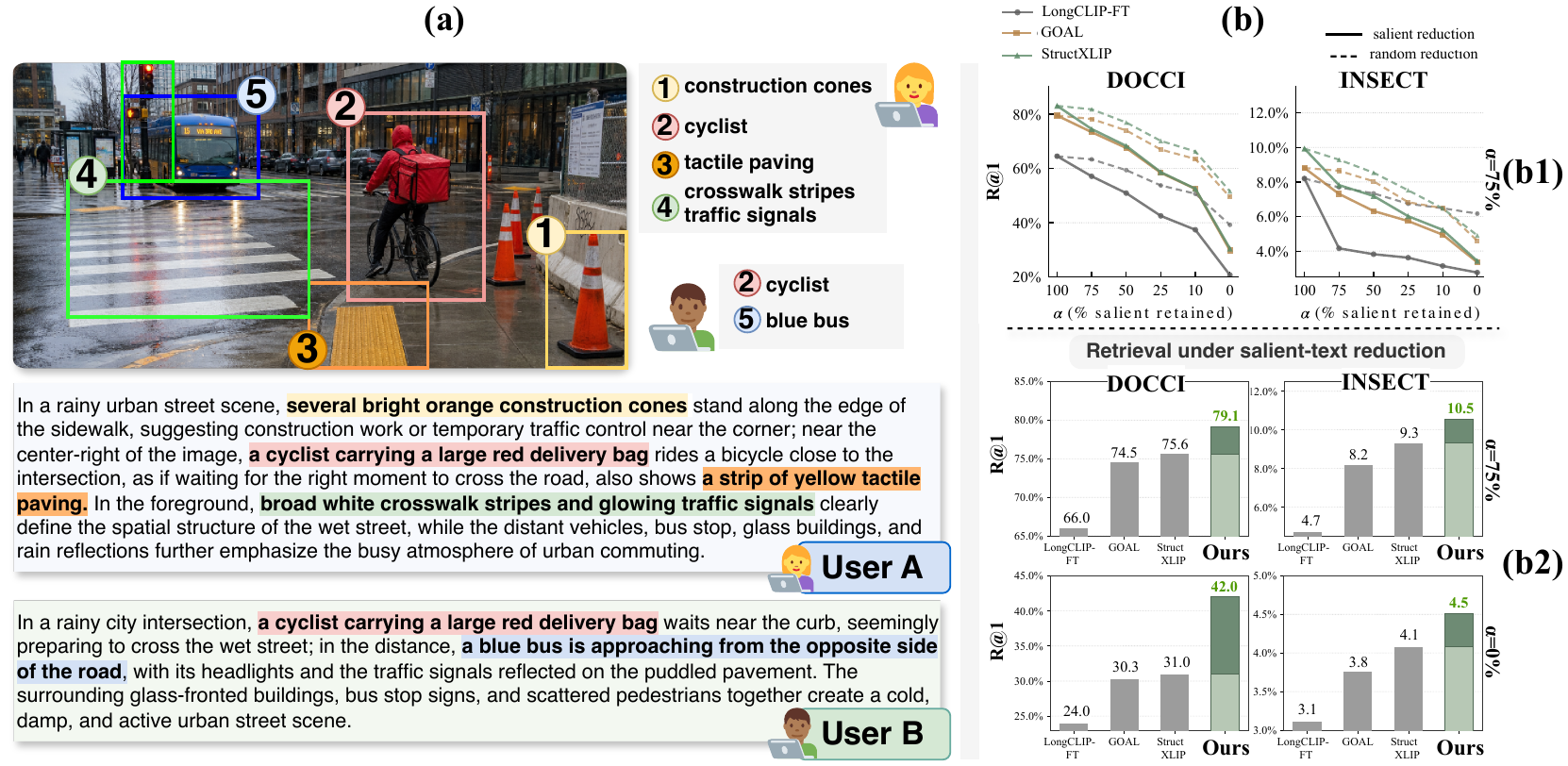}
    \vspace{-5pt}
    \caption{\textbf{Saliency collapse in long-text alignment.} a) Different users may describe the same image with long queries that cover different subsets of salient objects; b1) Existing long-text alignment methods suffer from \textbf{\textit{saliency collapse}}: they degrade sharply when salient text is progressively reduced, at a faster pace than when random content of the matched-length is removed. b2) Under both mild salient text reduction ($\alpha = 75\%$) and complete salient text removal ($\alpha = 0\%$), our method significantly improves over prior methods in terms of R@1. The relative gain of ours over the best competitor StructXLIP~\citep{ruan2026StructXLIP} is highlighted in a darker tone.}
    \label{fig:teaser}
    \vspace{-10pt}
\end{figure}

Departing from existing literature, we revisit long-text alignment through a human-centric lens: humans do not process complex visual scenes uniformly, but naturally attend to specific, salient regions~\citep{borji2013stands,kumain2025revisited}. Our preliminary analysis of text-to-image retrieval on long-text benchmarks~\citep{docci2024,truong2025insect}, as shown in Fig.~\ref{fig:teaser}(b1), reveals that state-of-the-art alignment methods~\citep{zhang2024long,goal2025,ruan2026StructXLIP} rely disproportionately on scattered text describing visual prominence, \ie \textit{salient text}, while overlooking the surrounding non-salient context, \ie \textit{contextual text}, despite its semantic value for retrieval (details in Sec.~\ref{sec:preliminary}). 
When we progressively reduce the ratio of salient text ($\alpha$) within a query, the retrieval performance (R@1) of existing models degrades significantly faster than when an equivalent amount of text is removed at random. 
We term this failure mode as \textbf{\textit{saliency collapse}}\rebuttal{: Information-dense salient evidence provides an attractive retrieval shortcut. Overreliance on this shortcut can prevent models from effectively using the remaining contextual evidence when salient cues are weakened. This }
exposes a critical limitation for downstream tasks such as text-to-image retrieval, as humans can subjectively prioritize different aspects of a scene~\citep{rahman2015visual, liang2018capvis, zangrossi2021visual}. 
As illustrated in Fig.~\ref{fig:teaser}(a), when asked to describe the same image, user A might focus on more prominent foreground objects than user B. 

A truly robust vision-language aligned model must go beyond simple token extension to represent both salient visual prominence and the complementary, non-salient context. To address saliency collapse in long-text alignment, we propose \textit{\ourmethodfull{} (\ourmethod)}, a novel finetuning framework that exploits auxiliary learning to \rebuttal{explicitly} reinforce both salient and contextual content alignment in long text. 
\ourmethod~decomposes the original image-text pairs into two complementary views: a \textit{salient view} of image-text pair featuring visually prominent regions, and a \textit{contextual view} of image-text pair featuring the remaining non-salient context. These decomposed views then serve for \textit{auxiliary alignment objectives}, together with the primary image-text alignment objective. Specifically, the auxiliary alignment objectives contain i) saliency-focused contrastive alignment that operates among the salient views, ii) context-preserving contrastive alignment that operates among the contextual views, and iii) visual anchoring that prevents severe visual representation drift. Importantly, the saliency-guided decomposition is a fine-tuning phase only, introducing no additional inference cost.

The efficacy of this paradigm hinges on how saliency is defined across modalities. While visual saliency arises from both low-level cues and high-level semantics, natural language primarily captures the latter and struggles to express perceptual saliency driven by non-semantic signals~\citep{song2014low,liang2018capvis}. We argue that object-level saliency sits at the intersection, providing a natural bridge between vision and language.
Building on advances in object saliency estimation~\citep{borji2019salientsurvey,dhara2025survey} and evidence that modern LLMs exhibit human-like semantic attention~\citep{chen2025explainable,lee2025video}, we adopt a modality-independent decomposition strategy. Specifically, we use an off-the-shelf object saliency estimator~\citep{yildirim2014fasa} for images, while independently we exploit a Large Language Model (LLM) to decompose the captions. This modality-independent strategy is intuitive, as cross-modal approaches rely on modality-aligned models, and is further supported by our ablation studies.
As shown in Fig.~\ref{fig:teaser}(b2), under both mild salient text reduction ($\alpha=75\%$) and complete removal ($\alpha=0\%$), \ourmethod is more robust to saliency collapse, achieving higher retrieval performance than existing methods.
Extensive experiments across various benchmarks show that \ourmethod significantly improves long-text retrieval. The auxiliary objectives are also plug-and-play, yielding consistent boosts to existing models.

To summarize, our main contributions are as follows:
\begin{itemize}[noitemsep,topsep=0pt]
    \item We identify and characterize \textbf{\textit{saliency collapse}}, an overlooked vulnerability in long-text alignment, where models over-rely on the presence of salient text.
    \rebuttal{\item We propose Robust Saliency-aware Alignment (ROSA), a novel finetuning framework that addresses saliency collapse through complementary supervision of salient and contextual image--text views.}
    \item We propose a modality-independent strategy for saliency-guided view decomposition using object saliency as a proxy to bridge vision and language.
    \item We propose auxiliary saliency-aware alignment objectives that can operate effectively standalone and consistently boost existing fine-tuning methods in a plug-and-play manner.
    \item We demonstrate \ourmethod achieves best performance on standard long-text retrieval benchmarks, and superior robustness when salient text is progressively reduced.
\end{itemize}

\section{Related Works}

\noindent\textbf{Long-Text Vision-Language Alignment.} Vision-language models such as CLIP~\citep{radford2021learning} and SigLIP~\citep{zhai2023siglip} have achieved strong image-text retrieval performance through contrastive or pairwise alignment. However, most existing methods rely on global image-text matching, treating the entire caption as a unified supervision signal. With the rise of densely captioned datasets such as DOCCI~\citep{docci2024} and ShareGPT4V~\citep{chen2024sharegpt4v}, long-text alignment has attracted increasing attention. Recent methods, including Long-CLIP~\citep{zhang2024long}, TULIP~\citep{najdenkoska2024tulip}, GOAL~\citep{goal2025}, StructXLIP~\citep{ruan2026StructXLIP}, improve long-text modeling by extending text encoders, introducing global-local interactions, or enhancing structural cues. 
Nevertheless, existing methods rarely consider that long captions are not semantically uniform. A caption may describe the main object, the surrounding scene, and incidental details at the same time.
When all words are treated as the same supervision, salient and contextual semantics can become entangled, causing the model to over-rely on prominent cues while underutilizing contextual information. \ourmethod therefore separates salient semantics from contextual descriptions to enable structured joint learning of both components.

\noindent\textbf{Visual Saliency Estimation and Saliency-Guided Learning.}
Visual saliency aims to identify the most visually prominent regions in an image. Early methods relied on hand-crafted low-level cues, but with the development of deep learning, data-driven models like DeepGaze~\citep{kummerer2025msdb} and SALICON~\citep{salicon2015}, as well as recent universal segmentation foundation models like SAM~\citep{kirillov2023segment} and SEEM~\citep{zou2023seem}, have provided strong semantic region priors for obtaining object-level and region-level visual masks.
Saliency has been widely used as weak guidance in vision-language learning, often through saliency maps, foreground masks, or visual prompts that steer models toward discriminative regions~\citep{Shtedritski0V23,SaliencyPrompt,gordeev2026saliency,gong2026saliency}. 
However, existing methods mainly exploit saliency to guide visual attention, grounding, or reasoning faithfulness on the visual side, while the textual supervision is usually kept as a global caption, answer, or class-level prompt. 
In contrast, \ourmethod uses saliency as a multimodal training prior: it builds saliency and contextual visual views, while splitting the long caption into corresponding salient and contextual text segments.

\noindent\textbf{Fine-Grained and Decoupled Vision-Language Alignment.}
\rebuttal{To overcome the limitations of global alignment, early image-text retrieval methods such as SCAN~\citep{lee2018stacked}, IMRAM~\citep{chen2020imram}, and NAAF~\citep{zhang2022negative} explored region–word interactions, iterative cross-modal matching, and matched/mismatched fragment modeling, respectively.
More recent approaches extend fine-grained alignment to pretrained vision-language models. }
FineLIP~\citep{asokan2025finelip} models token-level interactions, while RegionCLIP~\citep{zhong2022regionclip} and Alpha-CLIP~\citep{sun2024alpha} guide models toward local visual evidence using region annotations or visual prompts. Recent methods further incorporate auxiliary visual cues, such as edge maps, to learn more structured alignment signals~\citep{ruan2026StructXLIP}. 
Our work is also related to decoupled alignment. DAPT~\citep{zhang2025decouple} proposes to decouple image-text alignment for short-text prompt tuning, mainly addressing background interference under category-level supervision. However, DAPT is difficult to extend to our setting because its decoupling is built upon category-level prompts, whereas long-text alignment involves class-agnostic descriptions with densely entangled salient objects, attributes, relations, and contextual details. In contrast, \ourmethod decomposes multimodal inputs from both image and the caption side, and enforces alignment on both salient and contextual parts.

\section{Preliminary Study}
\label{sec:preliminary}
In long-text image retrieval, user queries can differ in what they describe from the semantically rich image.
We define \textit{salient text} as the part of textual descriptions referring to visually salient regions.
Specifically, we study how existing long-text image-text alignment models behave when such salient information is incomplete or missing.

\noindent \textbf{Experimental setup.}
We conduct the study on DOCCI~\citep{docci2024} and INSECT~\citep{truong2025insect}, covering general domain and specific domain, respectively.
We evaluate LongCLIP-FT~\citep{zhang2024long}, GOAL~\citep{goal2025}, and StructXLIP~\citep{ruan2026StructXLIP}, which respectively represent long-context encoding, local fine-grained alignment, and structured visual modeling.
All models are fine-tuned on each dataset's training split and evaluated on the corresponding test split.
During analysis, we keep model parameters fixed and perturb only the text queries.
We report text-to-image R@1 averaged over three independent runs.
For each caption, we leverage a LLM to partition the original text into two disjoint sets: $\texts$ contains visually salient descriptions, and $\textc$ contains the remaining contextual information. Note that this process is the same as in Method for the text decomposition (Sec.~\ref{sec:method}).
We can then perturb the text query by controlling the ratio of salient-text $\alpha$, defined as the percentage of tokens in $\texts$ kept in the perturbed query.

\noindent\colorbox{gray!10}{\textbf{Analysis 1: Do existing models mainly rely on salient text?}}
We first examine whether existing long-text alignment models rely primarily on $\texts$ for image-text matching. To this end, we focus on \emph{salient text} only and discard all non-salient context. For each caption, we keep only the first-$\alpha\%$ salient text tokens, where $\alpha \in \{100, 50, 25, 10\}$. 
As shown in Tab.~\ref{tab:salient_only}, 
$\texts$ alone ($\alpha{=}100$) retains more than 80\% of the corresponding full-caption R@1, even after all contextual text is removed.
As $\alpha$ decreases, R@1 drops monotonically across all models on both datasets, as can be observed by the performance ratio (superscript in gray) relative to full-caption retrieval.


\newcommand{\rr}[2]{#1\textsuperscript{\tiny\textcolor{black!55}{\,#2\%}}}

\begin{table}[t]
\centering
\caption{\textbf{Salient-text-only analysis.} For each caption we retain only the first-$\alpha\%$ salient text and remove all contextual text. Cells report text-to-image R@1 (\%); \emph{Full} denotes retrieval with the full caption paired with the image. Superscripts indicate the performance ratio relative to \emph{Full}.}
\label{tab:salient_only}
\vspace{0pt}
\footnotesize
\setlength{\tabcolsep}{4pt}
\resizebox{\columnwidth}{!}{%
\begin{tabular}{@{} l c *{4}{c} | c *{4}{c} @{}}
\toprule
\multirow{2}{*}{\textbf{Method}} &
\multicolumn{5}{c|}{\cellcolor{panelgray}\textbf{DOCCI}} &
\multicolumn{5}{c}{\cellcolor{panelgray}\textbf{INSECT}} \\
\cmidrule(lr){2-6}\cmidrule(lr){7-11}
& \emph{Full} & $\alpha{=}100$ & $\alpha{=}50$ & $\alpha{=}25$ & $\alpha{=}10$
& \emph{Full} & $\alpha{=}100$ & $\alpha{=}50$ & $\alpha{=}25$ & $\alpha{=}10$ \\
\midrule
Long-CLIP {\tiny [ECCV'24]}
  & 64.49 & \rr{62.02}{96} & \rr{48.67}{75} & \rr{35.49}{55} & \rr{28.24}{44}
  & 8.20  & \rr{7.41}{90}  & \rr{5.35}{65}  & \rr{4.26}{52}  & \rr{3.81}{46} \\
GOAL {\tiny [CVPR'25]}
  & 79.47 & \rr{66.40}{84} & \rr{53.11}{67} & \rr{39.46}{50} & \rr{31.10}{39}
  & 8.81  & \rr{7.73}{88}  & \rr{4.76}{54}  & \rr{3.60}{41}  & \rr{2.72}{31} \\
\textbf{StructXLIP} {\tiny [CVPR'26]}
  & 83.04 & \rr{66.98}{81} & \rr{53.20}{64} & \rr{38.75}{47} & \rr{30.40}{37}
  & 9.93  & \rr{8.00}{81}  & \rr{5.93}{60}  & \rr{3.90}{39}  & \rr{3.28}{33} \\
\bottomrule
\end{tabular}
}
\end{table}
\noindent\colorbox{gray!10}{\textbf{Finding 1.}}
Existing models strongly rely on salient text: salient text alone retains a substantial portion of full-caption retrieval performance, while progressively reducing it leads to consistent degradation.

\begin{figure}[t!]
    \centering
    \includegraphics[width=1.0\linewidth]{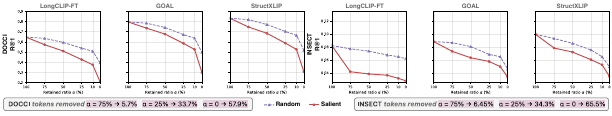}
    \vspace{-15pt}
\caption{
\textbf{Length-controlled analysis under progressive salient reduction.}
We compare retrieval performance under
\salientlegend~\textit{salient reduction} and
\randomlegend~\textit{random-span reduction}
on DOCCI and INSECT.
For each retained salient-text ratio $\alpha$, the random-span control removes the same number of tokens as salient deletion, isolating the effect of shortening from the effect of removing salient content.
The annotations report the corresponding percentage of the \emph{original caption} removed.
Across all panels, salient deletion yields much larger degradation than matched random deletion, indicating that the drop is caused by the absence of salient text rather than by reduced query length alone.
}
    \label{fig:lengthcontrol}
    \vspace{-10pt}
\end{figure}

\rebuttal{\noindent\colorbox{gray!10}{\textbf{Analysis 2:  Is the degradation caused by textual confounders?}}}
A natural concern is that the retrieval drop may result from shorter captions. To separate the effect of text length from the effect of removing salient content, we introduce a length-matched random control: for each salient-text ratio $\alpha$, we remove from the original caption a matched-length text at a random position.
Fig.~\ref{fig:lengthcontrol} compares salient reduction and random reduction on DOCCI and INSECT. We map $\alpha$ the ratio of salient text to the ratio of tokens of full-caption. For example, when $\alpha=75\%$, only 5.7\% of the original caption is removed on DOCCI and 6.45\% on INSECT.
Across models, curves of salient reduction decline more steeply than that of matched-length random control, and the gap widens as $\alpha$ decreases. 
\rebuttal{
Beyond length, we compare salient and contextual deletions from the same caption, matched separately by entity count, TF--IDF density, caption position, and E5-based text-only discriminability, to investigate whether these textual factors explain the greater sensitivity to salient-content removal. 
Averaged over deletion ratios, salient deletion yields lower R@1 under every control (\supp Sec.~\ref{sec:controlled_deletion}).
}


\rebuttal{
\noindent\colorbox{gray!10}{\textbf{Finding 2.}}
The loss of salient information leads to a larger average performance drop under separate controls for query length and other textual confounders, indicating concentrated reliance on salient evidence.
}

\noindent\colorbox{gray!10}{\textbf{Analysis 3: Is non-salient context informative for retrieval?}}
To investigate this, we employ E5-large-v2~\citep{wang2022e5}, a standalone text encoder independent of image–text alignment training, as a text-only oracle. Using only the non-salient context as the query, we attempt to retrieve the corresponding full caption from the caption database. Interestingly, this approach achieves 71.86\% R@1 on DOCCI and 41.90\% R@1 on INSECT.
\rebuttal{
Separately, for text-to-image retrieval using the same non-salient queries, the strongest baseline, StructXLIP~\citep{ruan2026StructXLIP}, attains only 30.96\% R@1 on DOCCI and 4.08\% R@1 on INSECT.
}
This substantial gap suggests that non-salient context is indeed informative, yet current multimodal alignment models fail to represent it effectively.

\noindent\colorbox{gray!10}{\textbf{Finding 3.}}
The non-salient context text is informative, however it is underutilized in existing long-text multi-modal aligned models.

\noindent\colorbox{blue!10}{\textbf{Summary.}}
These analyses reveal an interesting failure mode that we term as \textit{\textbf{saliency collapse}}: existing long-text image-text aligned models rely on salient text, and their retrieval accuracy drops disproportionately once such salient cues are incomplete.
Yet, long captions often contain heterogeneous visual semantics, featuring not only salient regions, but also rich contextual cues. The model should learn to encode both salient signals and the complementary context evidence.
\section{Method}
\label{sec:method}
\begin{figure}[t!] 
  \centering
  \includegraphics[width=0.9\linewidth]{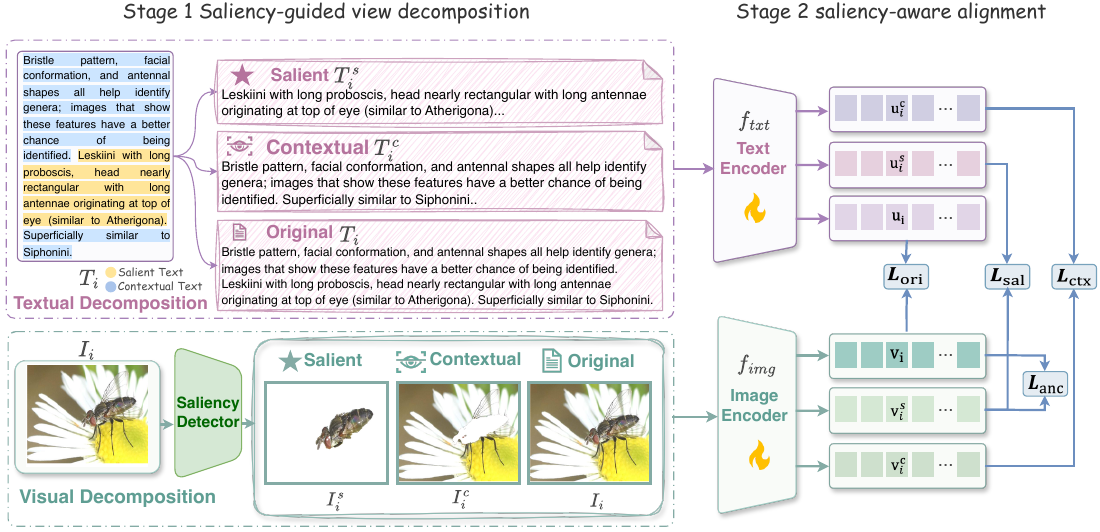} 
  \vspace{-5pt}
  \caption{\textbf{Overview of \ourmethod.}
  \ourmethod consists of two stages: \textit{\textbf{Stage 1: Saliency-guided view decomposition}} of vision and text, where we independently decompose the text and image. On the text side, an LLM is used to split the original long caption into salient text and contextual text. On the visual side, an off-the-shelf saliency method is applied to derive salient/contextual visual regions. \textit{\textbf{Stage 2: Saliency-aware alignment.}} During fine-tuning of the image and text encoders, \ourmethod augments the original image-text contrastive loss $\mathcal{L}{ori}$ with three saliency-aware objectives: the saliency-focused contrastive loss $\mathcal{L}{sal}$, which explicitly aligns salient images with salient texts; the context-preserving contrastive loss $\mathcal{L}{ctx}$, which aligns contextual images with contextual texts; and the visual anchoring loss $\mathcal{L}{anc}$, which enforces representation consistency between the original image and the salient image, thereby preserving stable salient semantics.
  }
  \vspace{-10pt}
  \label{fig:framework}
\end{figure}
Motivated by Sec.~\ref{sec:preliminary}, to address \textbf{\textit{saliency collapse}}, we propose \ourmethodfull{}, \ourmethod, a saliency-guided auxiliary learning framework that jointly learns the visual-textual correspondences of both salient and complementary contextual information.
As shown in Fig.~\ref{fig:framework}, \ourmethod decomposes each image-caption pair $(\image_i,\desc_i)$ into a salient view $(\images_i,\texts_i)$ and a contextual view $(\imagec_i,\textc_i)$. The former captures salient visual-textual cues, while the latter features contextual but informative visual-textual cues.
After decomposition, \ourmethod fine-tunes the visual and textual encoders by augmenting the original image-text contrastive loss $\mathcal{L}{ori}$ with three saliency-aware objectives.
The saliency-focused contrastive loss $\mathcal{L}{sal}$, as motivated by \textbf{Finding 1}, explicitly aligns salient images with salient texts to strengthen grounding between salient regions and salient text.
The context-preserving contrastive loss $\mathcal{L}{ctx}$, as motivated by \textbf{Finding 2\&3}, aligns contextual images with contextual texts to exploit informative but under-utilized contextual evidence. Finally, the visual anchoring loss $\anchorloss$ mitigates saliency-induced representation drift. Auxiliary views are used only during fine-tuning, so inference follows the standard CLIP pipeline with no extra cost.


\noindent \textbf{Problem formulation.} Let $\{(\image_i,\desc_i)\}_{i=1}^{N}$ denote a mini-batch of image-caption pairs, where $\image_i \in \visualspace$ is an image and $\desc_i \in \textspace$ is its corresponding long textual description. 
A vision-language model consists of an image encoder $\visualencoder$ and a text encoder $\textencoder$, which map the two modalities into a shared $d$-dimensional embedding space, \ie $\imagefeat_i = \visualencoder(\image_i)$, and $\textfeat_i = \textencoder(\desc_i)$. The standard image-text contrastive objective aligns the representation of $\image_i$ with that of $\desc_i$ while separating it from other samples in the batch. Let $s_{i,j}$ be the image-to-text similarity.
The original image-text alignment loss $\originalloss$ is expressed as the symmetric InfoNCE loss:
\begin{equation}
\originalloss =
\frac{1}{2N}
\sum_{i=1}^{N}
\left[
-\log
\frac{
    \exp(s_{i,i}/\temperature)
}{
    \sum_{j=1}^{N}\exp(s_{i,j}/\temperature)
}
-
\log
\frac{
    \exp(s_{i,i}/\temperature)
}{
    \sum_{j=1}^{N}\exp(s_{j,i}/\temperature)
}
\right],
\label{eq:original_loss}
\end{equation}
where $\temperature$ is the temperature parameter.

\noindent\textbf{Saliency-guided view decomposition.}
For each image-caption pair $(\image_i,\desc_i)$, we construct a salient pair $(\images_i,\texts_i)$ and a contextual pair $(\imagec_i,\textc_i)$ through modality-independent decomposition, where visual and textual views are obtained separately without cross-modal guidance, as detailed in \supp Sec.~\ref{sec:generation_validation}. 
A saliency detector produces $\saliencymap_i=\saliency(\image_i)$, which is converted into a mask $\saliencymask_i$ to form:
\begin{equation}
    \images_i = \maskop(\image_i,\saliencymask_i),
    \quad
    \imagec_i = \contextop(\image_i,\saliencymask_i),
\label{eq:visual_views}
\end{equation}
where $\images_i$ keeps visually prominent regions and $\imagec_i$ keeps the complementary context. 
Independently, an LLM partitions the caption with a fixed prompt $\prompt$:
\begin{equation}
    (\texts_i,\textc_i) = \partitioner(\desc_i,\prompt),
\label{eq:text_partition}
\end{equation}
where $\texts_i$ refers to salient visual content and $\textc_i$ to the remaining context. 
The LLM only partitions the input caption without rewriting or generating new content.
\rebuttal{
Invalid, unsupported, or incomplete outputs are automatically detected and regenerated.
}
We show the prompt $\prompt$ in \supp Fig.~\ref{fig:prompt}.
\rebuttal{
Human studies further validate the visual grounding of both salient and contextual text (\supp Tab.~\ref{tab:human_validation_sc}). 
}



\noindent\textbf{Saliency-focused contrastive alignment.} This objective aims to explicitly align salient visual regions with the salient text. For a mini-batch of salient view pairs $\{(\images_i,\texts_i)\}_{i=1}^{N}$, we compute their embeddings $\imagesfeat_i = \visualencoder(\images_i)$ and $\textsfeat_i = \textencoder(\texts_i)$ to obtain their image-text similarity $s^{s}_{i,j}$.
We then optimize the saliency-focused contrastive loss $\salignloss$:
\begin{equation}
\salignloss =
\frac{1}{2N}
\sum_{i=1}^{N}
\left[
-\log
\frac{
    \exp(s^{s}_{i,i}/\temperature)
}{
    \sum_{j=1}^{N}\exp(s^{s}_{i,j}/\temperature)
}
-
\log
\frac{
    \exp(s^{s}_{i,i}/\temperature)
}{
    \sum_{j=1}^{N}\exp(s^{s}_{j,i}/\temperature)
}
\right].
\label{eq:saliency_loss}
\end{equation}
$\salignloss$ strengthens the correspondence between salient visual regions and salient textual descriptions.
\noindent \textbf{Context-preserving contrastive alignment.}
We further align the contextual visual view $\imagec_i$ with the contextual text $\textc_i$ through a context-preserving contrastive objective.
Similar to the salient view pairs, for each contextual view pair $(\imagec_i,\textc_i)$, we compute their embeddings $\imagecfeat_i = \visualencoder(\imagec_i)$ and $\textcfeat_i = \textencoder(\textc_i)$ to then compute the contextual image-text similarity $s^{c}_{i,j}$.
The context-preserving contrastive loss $\contextloss$ is then expressed as:
\begin{equation}
\contextloss =
\frac{1}{2N}
\sum_{i=1}^{N}
\left[
-\log
\frac{
    \exp(s^{c}_{i,i}/\temperature)
}{
    \sum_{j=1}^{N}\exp(s^{c}_{i,j}/\temperature)
}
-
\log
\frac{
    \exp(s^{c}_{i,i}/\temperature)
}{
    \sum_{j=1}^{N}\exp(s^{c}_{j,i}/\temperature)
}
\right].
\label{eq:context_loss}
\end{equation}
$\contextloss$ encourages the learning of non-salient but informative cues to be more robust when salient information is incomplete.

\noindent \textbf{Visual anchoring.}
Since inference is performed using the original image $\image_i$, the saliency-guided fine-tuning should not cause the original image representation to drift away from the pre-trained vision-language embedding space. 
Following the intuition that the salient view $\images$ contains the most prominent visual content of the image, its representation should also be close to the one of the original image. We thus define the visual anchoring loss $\anchorloss$ as:
\begin{equation}
    \anchorloss
    =
    \frac{1}{N}
    \sum_{i=1}^{N}
    \left(
        1 -
        \frac{
            \visualencoder(\image_i)^{\top}
            \visualencoder(\images_i)
        }{
            \|\visualencoder(\image_i)\|_2
            \|\visualencoder(\images_i)\|_2
        }
    \right).
\label{eq:anchor_loss}
\end{equation}
$\anchorloss$ aims to prevent severe representation drift.

\noindent\textbf{Full Training Objective.}
The final objective combines the standard full-image full-caption contrastive loss with the proposed auxiliary losses:
\begin{equation}
\begin{aligned}
\totalloss
=
\originalloss
+
\lambdasal \salignloss
+
\lambdactx \contextloss
+
\lambdaanc \anchorloss ,
\end{aligned}
\label{eq:total_loss}
\end{equation}
where $\lambdasal$, $\lambdactx$, and $\lambdaanc$ control the contributions of the auxiliary objectives.
\section{Experiments}
\label{sec:experiments}

\textbf{Datasets and Metrics.}
Following prior evaluation protocols~\citep{ruan2026StructXLIP}, we evaluate our method on four fine-grained image-text retrieval benchmarks: DOCCI~\citep{docci2024} and DCI~\citep{urbanek2024picture} for general-domain dense caption retrieval, Sketchy~\citep{girella2025lots} for fashion-domain garment descriptions, and INSECT~\citep{truong2025insect} for expert-verified biological descriptions. 
These datasets cover diverse long-text alignment scenarios.
We report Recall@K (R@K, K=1/5/10) for both \inlineColorbox{lightpink}{\emph{Text$\rightarrow$Image(T$\rightarrow$I)}} and
\inlineColorbox{lightblue}{\emph{Image$\rightarrow$Text(I$\rightarrow$T)}} retrieval. For more details on the datasets refer to \supp Sec. ~\ref{sec:dataset_details}.

\textbf{Compared methods.}
Following the baseline selection of StructXLIP~\citep{ruan2026StructXLIP}, we compare \ourmethod with representative state-of-the-art CLIP-based fine-tuning methods for long-text image-text alignment, including Long-CLIP~\citep{zhang2024long}, FineLIP~\citep{asokan2025finelip}, SmartCLIP~\citep{smartclip2025}, GOAL~\citep{goal2025}, and StructXLIP~\citep{ruan2026StructXLIP}. These baselines cover major directions such as long-context modeling, fine-grained alignment, global-local correspondence, and multimodal structural cue learning.

\textbf{Implementation details.}
We follow the experimental setting of~\citep{ruan2026StructXLIP} for model configuration and training, using CLIP ViT-B/16 as the visual encoder and Knowledge Preserving Stretching(KPS)~\citep{zhang2024long} to extend the text encoder, while keeping all learning rates and optimizer configurations unchanged. For saliency-guided decomposition, we use~\citep{yildirim2014fasa} to extract salient visual regions, and obtain the contextual view by removing the salient region from the original image. On the textual side, we use Qwen2.5~\citep{qwen2025qwen25technicalreport} to extract salient and contextual texts from each caption. Images are resized to $224 \times 224$, and all models are fine-tuned for 10 epochs on NVIDIA RTX 5090 GPUs with 32 GB memory and NVIDIA A100 GPU with 40 GB memory. Each run requires approximately 1-3 hours, depending on the dataset. We report the average results over three random seeds. 
\rebuttal{For a direct training-cost comparison, StructXLIP requires 0.17,s per batch, while \ourmethod requires 0.34,s per batch under the same training setting. The additional cost comes from encoding the salient and contextual views during fine-tuning; preprocessing is performed once offline and cached, and \ourmethod introduces no additional parameters or inference overhead.}

\subsection{Comparison with existing CLIP-based methods}


\begin{table}
\centering
\caption{
\textbf{Comparison against recent CLIP-based finetuning methods on long-text cross-modal retrieval}. We report Recall@K (\%) on both \inlineColorbox{lightpink}{\emph{T$\rightarrow$I}} and \inlineColorbox{lightblue}{\emph{I$\rightarrow$T}} settings and highlight \textbf{best} and \underline{second best} results; $\Delta$ denotes the margin of ours over the best competitor.
}
\label{tab:main_results}
\footnotesize
\setlength{\tabcolsep}{2pt} 
\resizebox{\textwidth}{!}{
\begin{tabular}{@{} l *{6}{c} | *{6}{c} | *{6}{c}| *{6}{c} @{}}
\toprule

\multirow{2}{*}{\textbf{Method}} &
\multicolumn{6}{c|}{\cellcolor{panelgray}\textbf{SKETCHY}} &
\multicolumn{6}{c|}{\cellcolor{panelgray}\textbf{INSECT}}  &
\multicolumn{6}{c|}{\cellcolor{panelgray}\textbf{DOCCI}} &
\multicolumn{6}{c}{\cellcolor{panelgray}\textbf{DCI}} \\
\cmidrule(lr){2-7}\cmidrule(lr){8-13}\cmidrule(lr){14-19}\cmidrule(lr){20-25}

& \cellcolor{lightpink}{R@1} & \cellcolor{lightpink}{R@5} & \cellcolor{lightpink}{R@10}
& \cellcolor{lightblue}{R@1} & \cellcolor{lightblue}{R@5} & \multicolumn{1}{c|}{\cellcolor{lightblue}{R@10}}
& \cellcolor{lightpink}{R@1} & \cellcolor{lightpink}{R@5} & \cellcolor{lightpink}{R@10}
& \cellcolor{lightblue}{R@1} & \cellcolor{lightblue}{R@5} & \multicolumn{1}{c|}{\cellcolor{lightblue}{R@10}}
& \cellcolor{lightpink}{R@1} & \cellcolor{lightpink}{R@5} & \cellcolor{lightpink}{R@10}
& \cellcolor{lightblue}{R@1} & \cellcolor{lightblue}{R@5} & \multicolumn{1}{c|}{\cellcolor{lightblue}{R@10}}
& \cellcolor{lightpink}{R@1} & \cellcolor{lightpink}{R@5} & \cellcolor{lightpink}{R@10}
& \cellcolor{lightblue}{R@1} & \cellcolor{lightblue}{R@5} & \cellcolor{lightblue}{R@10}\\

\midrule
Long-CLIP {\tiny [ECCV'24]} 
& 54.32 & 80.14 & 88.43 & 52.76 & 80.31 & 88.08 
& 8.20 & 23.83 & 34.97 & 9.41 & 24.78 & 37.31 
& 64.49 & 87.67 & 93.43 & 63.08 & 87.45 & 93.14 
& 59.23 & 80.89 & 87.04 & 60.13 & 81.44 & 87.54 \\

FineLIP {\tiny [CVPR'25]} 
& 40.59 & 71.16 & 81.78 & 40.33 & 72.11 & 82.38 
& 8.46 & 23.32 & 33.59 & 6.86 & 23.75 & 34.46 
& 67.80 & 90.22 & 94.84 & 66.39 & 89.12 & 94.47 
& 66.13 & 85.34 & 89.79 & 64.58 & 84.59 & 89.54 \\

SmartCLIP {\tiny [CVPR'25]} 
& 50.73 & 81.09 & 94.56 & 51.30 & 80.83 & 94.04 
& 4.84 & 16.84 & 34.63 & 4.66 & 15.46 & 34.02 
& 74.92 & 94.08 & 97.31 & 74.91 & 94.04 & 97.29 
& 69.88 & 86.64 & 94.05 & 70.94 & 87.04 & 92.77 \\

GOAL {\tiny [CVPR'25]}  
& 63.21 & 87.13 & 93.44 & 62.44 & 87.82 & 92.31 
& 8.81 & 24.35 & 35.84 & 8.55 & 25.91 & 36.18 
& 79.47 & 96.65 & 98.69 & 79.43 & 96.14 & 97.25 
& 72.64 &  89.89  & 93.70 &  72.84 &  \underline{90.50} &  93.20 \\

StructXLIP{\tiny [CVPR'26]} 
& \underline{69.86} & \underline{90.85} & \underline{95.42} & \underline{68.22} & \underline{90.67} & \underline{95.68} 
& \underline{9.93} & \underline{26.60} & \underline{38.34} & \underline{9.50} & \underline{26.60} & \underline{39.64} 
& \underline{83.04} & \underline{97.06} & \underline{98.96} & \underline{81.59} & \underline{96.94} & \underline{98.78} 
& \underline{75.90} & \underline{90.00} & \textbf{95.15} & \underline{74.39} & 89.90 & \textbf{94.30} \\
\midrule

\textbf{\ourmethod (Ours) }
& \textbf{71.5}& \textbf{92.31} & \textbf{95.85} &\textbf{71.5} & \textbf{91.62}& \textbf{95.85}
& \textbf{10.36} & \textbf{28.32} & \textbf{40.67} & \textbf{9.84} & \textbf{29.36} & \textbf{40.41} 
& \textbf{84.35}& \textbf{97.51}& \textbf{99.18}&\textbf{82.18} & \textbf{97.14} & \textbf{98.94}
& \textbf{77.09}&	\textbf{90.55}&	\underline{94.25}&	\textbf{74.99}&	\textbf{90.55}&	\underline{94.20} \\

\midrule

\textit{$\Delta$} 
& \gain{1.64} & \gain{1.46} & \gain{0.43} & \gain{3.28} & \gain{0.95} & \gain{0.17} 
& \gain{0.43} & \gain{1.72} & \gain{2.33} & \gain{0.34} & \gain{2.76} & \gain{0.77} 
& \gain{1.31} & \gain{0.45} & \gain{0.22} & \gain{0.59} & \gain{0.2} & \gain{0.16} 
& \gain{1.19} & \gain{0.55} & \down{0.9} & \gain{0.6} & \gain{0.05} & \down{0.1}  \\
\bottomrule
\end{tabular}
}
\vspace{-15pt}
\end{table}

\begin{table}[t]
\caption{
\textbf{Plug-and-play enhancement with our saliency-aware alignment objectives.} 
Results on \inlineColorbox{lightpink}{\emph{T$\rightarrow$I}} and 
\inlineColorbox{lightblue}{\emph{I$\rightarrow$T}} retrieval for INSECT and DOCCI, with \textbf{best} highlighted and the margin in bracket.}
\label{tab:plug-in}
\centering
\footnotesize
\setlength{\tabcolsep}{3pt}
\resizebox{\columnwidth}{!}{%
\begin{tabular}{@{} l*{5}{c!{\color{lightlightgray}\vrule width 0.5pt}} c | *{5}{c!{\color{lightlightgray}\vrule width 0.5pt}} c @{}}
\toprule
\multirow{2}{*}{\textbf{Method}} &
\multicolumn{6}{c|}{\cellcolor{gray!15}\textbf{INSECT}} &
\multicolumn{6}{c}{\cellcolor{gray!15}\textbf{DOCCI}} \\
\cmidrule(lr){2-7}\cmidrule(lr){8-13}
& \cellcolor{lightpink}{R@1} & \cellcolor{lightpink}{R@5} & \cellcolor{lightpink}{R@10}
& \cellcolor{lightblue}{R@1} & \cellcolor{lightblue}{R@5} & \multicolumn{1}{c|}{\cellcolor{lightblue}{R@10}}
& \cellcolor{lightpink}{R@1} & \cellcolor{lightpink}{R@5} & \cellcolor{lightpink}{R@10}
& \cellcolor{lightblue}{R@1} & \cellcolor{lightblue}{R@5} & \cellcolor{lightblue}{R@10} \\
\midrule

Long-CLIP & 8.20 & 23.83 & 34.97 & 9.41 & 24.78 & 37.31 & 64.49 & 87.67 & 93.43 & 63.08 & 87.45 & 93.14 \\
\textbf{+our $\ourloss$} 
& \textbf{9.41}\smallgain{1.21} 
& \textbf{25.73}\smallgain{1.90} 
& \textbf{37.82}\smallgain{2.85} 
& \textbf{10.36}\smallgain{0.95} 
& \textbf{28.58}\smallgain{3.80} 
& \textbf{40.07}\smallgain{2.76} 
& \textbf{67.73}\smallgain{3.24} 
& \textbf{90.63}\smallgain{2.96} 
& \textbf{94.98}\smallgain{1.55} 
& \textbf{67.29}\smallgain{4.21} 
& \textbf{89.37}\smallgain{1.92} 
& \textbf{94.39}\smallgain{1.25} \\
\midrule

FineLIP & 8.46 & 23.32 & 33.59 & 6.86 & 23.75 & 34.46 & 67.80 & 90.22 & 94.84 & 66.39 & 89.12 & 94.47 \\
\textbf{+our $\ourloss$} 
& \textbf{8.89}\smallgain{0.43} 
& \textbf{24.18}\smallgain{0.86} 
& \textbf{34.97}\smallgain{1.38} 
& \textbf{8.12}\smallgain{1.26} 
& \textbf{24.96}\smallgain{1.21} 
& \textbf{35.58}\smallgain{1.12} 
& \textbf{74.06}\smallgain{6.26} 
& \textbf{93.86}\smallgain{3.64} 
& \textbf{97.02}\smallgain{2.18} 
& \textbf{72.71}\smallgain{6.32} 
& \textbf{92.90}\smallgain{3.78} 
& \textbf{96.90}\smallgain{2.43} \\
\midrule

SmartCLIP & 4.84 & 16.84 & 34.63 & 4.66 & 15.46 & 34.02 & 74.92 & 94.08 & 97.31 & 74.91 & 94.04 & 97.29 \\
\textbf{+our $\ourloss$} 
& \textbf{8.46}\smallgain{3.62} 
& \textbf{24.78}\smallgain{7.94} 
& \textbf{35.06}\smallgain{0.43} 
& \textbf{7.69}\smallgain{3.03} 
& \textbf{24.61}\smallgain{9.15} 
& \textbf{35.84}\smallgain{1.82} 
& \textbf{80.61}\smallgain{5.69} 
& \textbf{96.63}\smallgain{2.55} 
& \textbf{98.69}\smallgain{1.38} 
& \textbf{78.59}\smallgain{3.68} 
& \textbf{95.96}\smallgain{1.92} 
& \textbf{98.18}\smallgain{0.89} \\
\midrule

GOAL & 8.81 & 24.35 & 35.84 & 8.55 & 25.91 & 36.18 & 79.47 & 96.65 & 98.69 & 79.43 & 96.14 & 98.51 \\
\textbf{+our $\ourloss$} 
& \textbf{9.67}\smallgain{0.86} 
& \textbf{27.55}\smallgain{3.20} 
& \textbf{39.72}\smallgain{3.88} 
& \textbf{8.72}\smallgain{0.17} 
& \textbf{27.89}\smallgain{1.98} 
& \textbf{40.16}\smallgain{3.98} 
& \textbf{83.86}\smallgain{4.39} 
& \textbf{97.47}\smallgain{0.82} 
& \textbf{99.16}\smallgain{0.47} 
& \textbf{82.02}\smallgain{2.59} 
& \textbf{97.27}\smallgain{1.13} 
& \textbf{98.96}\smallgain{0.45} \\
\midrule

SigLIP2 & 6.37 & 21.07 & \textbf{31.87} & 6.56 & 21.85 & 31.95 & 71.80 & 92.53 & 95.88 & 71.51 & 92.41 & 96.06 \\
\textbf{+our $\ourloss$} 
& \textbf{6.74}\smallgain{0.37} 
& \textbf{22.11}\smallgain{1.04} 
& 31.78\smalldown{0.09} 
& \textbf{7.43}\smallgain{0.87} 
& \textbf{22.54}\smallgain{0.69} 
& \textbf{32.82}\smallgain{0.87} 
& \textbf{80.12}\smallgain{8.32} 
& \textbf{96.27}\smallgain{3.74} 
& \textbf{98.61}\smallgain{2.73} 
& \textbf{78.73}\smallgain{7.22} 
& \textbf{96.25}\smallgain{3.84} 
& \textbf{98.37}\smallgain{2.31} \\
\midrule[1.5pt]

LoRA & 5.79 & 16.84 & 25.82 & 4.84 & 16.49 & 24.96 & 77.80 & 96.45 & 98.55 & 77.00 & 96.02 & 98.05 \\
\textbf{+our $\ourloss$} 
& \textbf{6.04}\smallgain{0.25} 
& \textbf{18.83}\smallgain{1.99} 
& \textbf{27.29}\smallgain{1.47} 
& \textbf{5.70}\smallgain{0.86} 
& \textbf{18.74}\smallgain{2.25} 
& \textbf{27.37}\smallgain{2.41} 
& \textbf{80.49}\smallgain{2.69} 
& \textbf{96.63}\smallgain{0.18} 
& \textbf{98.78}\smallgain{0.23} 
& \textbf{79.94}\smallgain{2.94} 
& \textbf{96.65}\smallgain{0.63} 
& \textbf{98.49}\smallgain{0.44} \\
\midrule

DoRA & 7.17 & 20.21 & 29.88 & 6.04 & 20.81 & 31.52 & 66.27 & 90.50 & 95.18 & 65.65 & 89.55 & 94.78 \\
\textbf{+our $\ourloss$} 
& \textbf{9.41}\smallgain{2.24} 
& \textbf{26.25}\smallgain{6.04} 
& \textbf{36.96}\smallgain{7.08} 
& \textbf{9.84}\smallgain{3.80} 
& \textbf{25.82}\smallgain{5.01} 
& \textbf{37.13}\smallgain{5.61} 
& \textbf{72.73}\smallgain{6.46} 
& \textbf{93.61}\smallgain{3.11} 
& \textbf{97.39}\smallgain{2.21} 
& \textbf{72.24}\smallgain{6.59} 
& \textbf{93.71}\smallgain{4.16} 
& \textbf{96.98}\smallgain{2.20} \\
\midrule

StructXLIP & 9.93 & 26.60 & 38.34 & 9.50 & 26.60 & 39.64 & 83.04 & 97.06 & 98.96 & 81.59 & 96.94 & 98.78 \\
\textbf{+our $\ourloss$} 
& \textbf{10.36}\smallgain{0.43} 
& \textbf{30.92}\smallgain{4.32} 
& \textbf{42.66}\smallgain{4.32} 
& \textbf{9.84}\smallgain{0.34} 
& \textbf{30.83}\smallgain{4.23} 
& \textbf{42.40}\smallgain{2.76} 
& \textbf{84.20}\smallgain{1.16} 
& \textbf{97.78}\smallgain{0.72} 
& \textbf{99.08}\smallgain{0.12} 
& \textbf{82.69}\smallgain{1.10} 
& \textbf{97.37}\smallgain{0.43} 
& \textbf{98.98}\smallgain{0.20} \\

\bottomrule
\end{tabular}
}
\vspace{-15pt}
\end{table}
\noindent\textbf{Quantitative Results.} As shown in Tab.~\ref{tab:main_results}, 
\rebuttal{under standard full-caption retrieval,}
\rebuttal{\ourmethod achieves the best results on most metrics among all compared methods, on four cross-modal retrieval benchmarks, in both T$\rightarrow$I and I$\rightarrow$T directions.}
Especially on Sketchy dataset, \ourmethod improves R@1 by +1.64\% for T$\rightarrow$I retrieval and +3.28\% for I$\rightarrow$T retrieval over the best competitor. 
On the rare-domain dataset INSECT, all methods yield relatively low recall scores due to the limited coverage of such domains in VLM pre-training data. Nevertheless, our method still achieves consistent gains in both retrieval directions. 
On the general-domain DOCCI benchmark, our method also shows consistent improvements. This indicates that salient visual-textual components are effective for dense image-caption pairs that contain multiple entities, attributes, and fine-grained local descriptions.

\textbf{Plug-and-play enhancement.} To further validate the tuning-agnostic effectiveness of our proposed saliency-aware alignment objectives, we integrate our proposed losses $\ourloss$ into diverse CLIP-based finetuning methods, including PEFT techniques such as LoRA~\citep{hu2022lora} and DoRA~\citep{liu2024dora}, as well as state-of-the-art CLIP-based finetuning methods from the main comparison.
As shown in Tab.~\ref{tab:plug-in}, adding $\ourloss$ to these methods leads to consistent improvements, with the largest gains observed for DoRA with +7.08\% on INSECT T$\rightarrow$I R@10 and SigLIP2 with +8.32\% on DOCCI T$\rightarrow$I R@1. 
\rebuttal{Saliency-aware alignment objectives provide complementary supervision by jointly optimizing salient and contextual evidence, thereby reducing their utilization imbalance while serving as an effective plug-and-play enhancement across CLIP-based retrieval models.}

\textbf{Robustness under reduced salient text.} 
We evaluate retrieval robustness by constructing three Text$\rightarrow$Image query settings:\inlineColorbox{salientonly}{\emph{Salient-only}} that uses only salient text as query,
\inlineColorbox{salienthalf}{\emph{Salient$_{50\%}$+Context}} uses 50\% salient text mixed with context as query, and
\inlineColorbox{salientzero}{\emph{Context-only}} with only context text as query. 
Tab.~\ref{tab:salient_robustness} shows that \ourmethod achieves the best results in most settings across DOCCI, DCI, and INSECT. The gains are modest under \emph{Salient-only} queries, where object-centric cues are already sufficient for all methods. In contrast, the advantage becomes larger as salient cues are weakened or absent. On DOCCI, \ourmethod improves R@1 by 5.13\% under \emph{Salient$_{50\%}$+Context}, and by 11.04\% under \emph{Context-only}. Consistent gains on DCI and INSECT further show \ourmethod can effectively encode both salient and contextual semantics, being more robust to queries of different saliency levels. 
\rebuttal{To further evaluate sensitivity to query construction, we evaluate fixed checkpoints using test queries reconstructed with a rewritten prompt, Llama-3.1, or human annotations.  \ourmethod improves Context-only R@1 over StructXLIP under all evaluated alternatives (Supp. Sec.~\ref{sec:query_robustness}).}

\begin{figure}[htbp] 
    \vspace{-8pt}
  \centering
  \includegraphics[width=\linewidth]{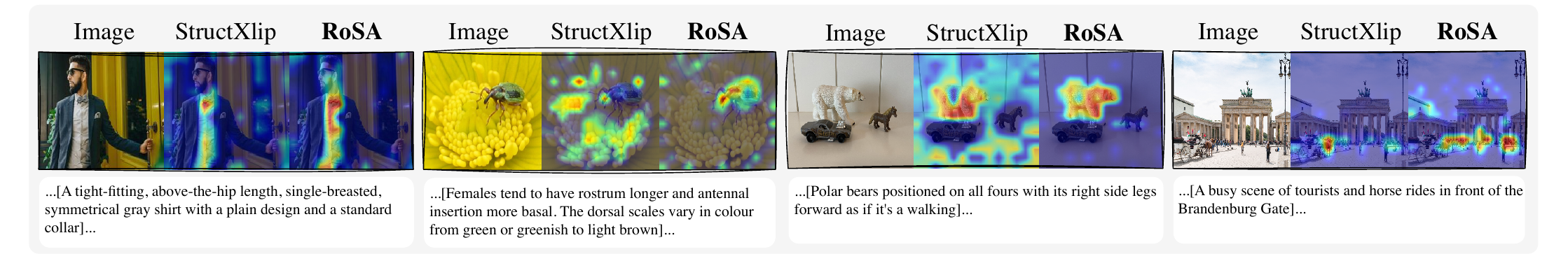} 
  \vspace{-15pt}
  \caption{Comparison of cross-modal Grad-CAM visualizations between StructXlip and \ourmethod.
From left to right, the columns show examples from Sketchy, INSECT, DCI, and DOCCI.}
\vspace{-10pt}
  \label{fig:main_vis}
\end{figure}

\textbf{Qualitative analysis.} We visualize the cross-modal attention maps of the competitor StructXLIP and \ourmethod on four datasets in Fig.~\ref{fig:main_vis}. \ourmethod shows a cleaner concentration on caption-related visual entities compared with StructXLIP, indicating that \ourmethod effectively represents relevant visual evidence at object level. Please refer to \supp Sec.~\ref{sec:qualitative} for more qualitative results, and Sec.~\ref{sec:pathgen} for other exploratory evaluation on medical domain.

\subsection{Ablation studies}

\begin{table}[t]
\centering
\caption{
\textbf{Robustness under reduced salient text.}
We report \inlineColorbox{lightpink}{\emph{T$\rightarrow$I}} Recall@K (\%) on DOCCI, DCI, and INSECT under reduced salient text:
\inlineColorbox{salientonly}{\emph{Salient-only}},
\inlineColorbox{salienthalf}{\emph{Salient$_{50\%}$+Context}}, and
\inlineColorbox{salientzero}{\emph{Context-only}}.
We highlight \textbf{best} and \underline{second best} results; $\Delta$ is our margin over the best competitor.
}
\label{tab:salient_robustness}
\scriptsize
\setlength{\tabcolsep}{1.5pt}
\renewcommand{\arraystretch}{1.05}
\resizebox{\textwidth}{!}{
\begin{tabular}{@{} l *{9}{c} | *{9}{c} | *{9}{c} @{}}
\toprule

\multirow{3}{*}{\textbf{Method}} &
\multicolumn{9}{c|}{\cellcolor{panelgray}\textbf{DOCCI}} &
\multicolumn{9}{c|}{\cellcolor{panelgray}\textbf{DCI}} &
\multicolumn{9}{c}{\cellcolor{panelgray}\textbf{INSECT}} \\

\cmidrule(lr){2-10}
\cmidrule(lr){11-19}
\cmidrule(lr){20-28}

& \multicolumn{3}{c}{\cellcolor{salientonly}{\textbf{Salient-only}}}
& \multicolumn{3}{c}{\cellcolor{salienthalf}{\textbf{Salient$_{50\%}$+Context}}}
& \multicolumn{3}{c|}{\cellcolor{salientzero}{\textbf{Context-only}}}

& \multicolumn{3}{c}{\cellcolor{salientonly}{\textbf{Salient-only}}}
& \multicolumn{3}{c}{\cellcolor{salienthalf}{\textbf{Salient$_{50\%}$+Context}}}
& \multicolumn{3}{c|}{\cellcolor{salientzero}{\textbf{Context-only}}}

& \multicolumn{3}{c}{\cellcolor{salientonly}{\textbf{Salient-only}}}
& \multicolumn{3}{c}{\cellcolor{salienthalf}{\textbf{Salient$_{50\%}$+Context}}}
& \multicolumn{3}{c}{\cellcolor{salientzero}{\textbf{Context-only}}} \\

\cmidrule(lr){2-4}
\cmidrule(lr){5-7}
\cmidrule(lr){8-10}
\cmidrule(lr){11-13}
\cmidrule(lr){14-16}
\cmidrule(lr){17-19}
\cmidrule(lr){20-22}
\cmidrule(lr){23-25}
\cmidrule(lr){26-28}

& \cellcolor{salientonly}{R@1}
& \cellcolor{salientonly}{R@5}
& \cellcolor{salientonly}{R@10}
& \cellcolor{salienthalf}{R@1}
& \cellcolor{salienthalf}{R@5}
& \cellcolor{salienthalf}{R@10}
& \cellcolor{salientzero}{R@1}
& \cellcolor{salientzero}{R@5}
& \multicolumn{1}{c|}{\cellcolor{salientzero}{R@10}}

& \cellcolor{salientonly}{R@1}
& \cellcolor{salientonly}{R@5}
& \cellcolor{salientonly}{R@10}
& \cellcolor{salienthalf}{R@1}
& \cellcolor{salienthalf}{R@5}
& \cellcolor{salienthalf}{R@10}
& \cellcolor{salientzero}{R@1}
& \cellcolor{salientzero}{R@5}
& \multicolumn{1}{c|}{\cellcolor{salientzero}{R@10}}

& \cellcolor{salientonly}{R@1}
& \cellcolor{salientonly}{R@5}
& \cellcolor{salientonly}{R@10}
& \cellcolor{salienthalf}{R@1}
& \cellcolor{salienthalf}{R@5}
& \cellcolor{salienthalf}{R@10}
& \cellcolor{salientzero}{R@1}
& \cellcolor{salientzero}{R@5}
& \cellcolor{salientzero}{R@10} \\

\midrule

Long-CLIP {\tiny [ECCV'24]}
& 61.54 & 84.71 & 90.82
& 58.87 & 83.35 & 89.52
& 23.96 & 43.90 & 52.34

& 51.64 & 71.73 & 80.15
& 54.01 & 77.00 & 84.07
& 24.64 & 45.12 & 53.43

& 6.55 & 20.41 & 31.04
& 4.30 & 17.11 & 24.88
& 3.11 & 11.60 & 17.40 \\

FineLIP {\tiny [CVPR'25]}
& 52.62 & 78.30 & 85.48
& 52.30 & 78.50 & 86.43
& 20.59 & 39.03 & 48.13

& 53.03 & 74.67 & 82.23
& 60.36 & 81.27 & 86.96
& 25.91 & 44.49 & 53.26

& \textbf{8.17} & 22.45 & 30.61
& 7.52 & 19.08 & 30.36
& 3.11 & 10.74 & 16.76 \\

SmartCLIP {\tiny [CVPR'25]}
& 61.88 & 85.48 & 91.69
& 54.17 & 81.36 & 88.84
& 24.92 & 45.01 & 53.30

& 57.24 & 77.38 & 84.25
& 65.03 & 84.19 & 88.82
& 30.01 & 49.45 & 57.24

& 2.36 & 7.20 & 10.74
& 2.18 & 7.41 & 11.71
& 1.40 & 4.83 & 7.09 \\

GOAL {\tiny [CVPR'25]}
& 66.84 & \underline{88.32} & \underline{93.21}
& 68.60 & \underline{90.85} & \underline{95.00}
& 30.28 & \underline{51.63} & \underline{60.09}

& 60.88 & 80.44 & \underline{85.92}
& 67.55 & 86.56 & 91.50
& 32.08 & 52.28 & \underline{59.67}

& 7.09 & 21.59 & 32.22
& 7.09 & 22.45 & \underline{32.72}
& 3.76 & \underline{11.92} & 17.51 \\

StructXLIP {\tiny [CVPR'26]}
& \underline{67.22} & 88.23 & 92.87
& \underline{69.31} & 90.75 & 94.90
& \underline{30.96} & 51.04 & 59.73

& \underline{61.97} & \underline{80.50} & 85.40
& \underline{70.65} & \underline{86.94} & \underline{91.52}
& \underline{34.28} & \underline{52.51} & 58.74

& \underline{8.16} & \underline{24.49} & \underline{35.12}
& \underline{8.56} & \underline{23.24} & 32.15
& \underline{4.08} & 11.49 & \underline{18.26} \\

\midrule

\textbf{\ourmethod(Ours)}
& \textbf{68.03} & \textbf{88.83} & \textbf{93.51}
& \textbf{74.44} & \textbf{93.82} & \textbf{97.01}
& \textbf{42.00} & \textbf{66.07} & \textbf{73.65}

& \textbf{62.26} & \textbf{80.84} & \textbf{86.21}
& \textbf{72.01} & \textbf{88.48} & \textbf{92.96}
& \textbf{37.56} & \textbf{57.76} & \textbf{65.20}

& \underline{8.16} & \textbf{25.89} & \textbf{37.49}
& \textbf{9.10} & \textbf{24.38} & \textbf{34.91}
& \textbf{4.88} & \textbf{14.82} & \textbf{22.34} \\

\midrule

\textit{$\Delta$}
& \gain{0.81} & \gain{0.51} & \gain{0.30}
& \gain{5.13} & \gain{2.97} & \gain{2.01}
& \gain{11.04} & \gain{14.44} & \gain{13.56}

& \gain{0.29} & \gain{0.34} & \gain{0.29}
& \gain{1.36} & \gain{1.54} & \gain{1.44}
& \gain{3.28} & \gain{5.25} & \gain{5.53}

& \down{0.01} & \gain{1.40} & \gain{2.37}
& \gain{0.54} & \gain{1.14} & \gain{2.19}
& \gain{0.80} & \gain{2.90} & \gain{4.08} \\

\bottomrule
\end{tabular}
}
\vspace{-10pt}
\end{table}

\begin{table*}[t]
\centering
\caption{
\textbf{Ablation on auxiliary saliency-aware alignment objectives.}
We report full-caption retrieval on \inlineColorbox{lightpink}{\emph{T$\rightarrow$I}} and
\inlineColorbox{lightblue}{\emph{I$\rightarrow$T}}, together in \inlineColorbox{lightpink}{\textit{T$\rightarrow$I robustness settings}} with reduced salient text including
\inlineColorbox{salientonly}{\emph{Salient-only}},
\inlineColorbox{salienthalf}{\emph{Salient$_{50\%}$+Context}}, and
\inlineColorbox{salientzero}{\emph{Context-only}}.
Rows ablate different combinations of $\salignloss$, $\contextloss$, and $\anchorloss$. We report Recall@K (\%) with \textbf{best} highlighted.
}
\vspace{0pt}
\label{tab:ablation_loss}
\scriptsize
\setlength{\tabcolsep}{2.0pt}
\renewcommand{\arraystretch}{1.12}
\resizebox{\textwidth}{!}{%
\begin{tabular}{@{} ccc | *{15}{c} | *{15}{c} @{}}
\toprule

\multirow{4}{*}{$\salignloss$}
& \multirow{4}{*}{$\contextloss$}
& \multirow{4}{*}{$\anchorloss$}
& \multicolumn{15}{c|}{\cellcolor{gray!15}\textbf{INSECT}}
& \multicolumn{15}{c}{\cellcolor{gray!15}\textbf{DOCCI}} \\

\cmidrule(lr){4-18}
\cmidrule(lr){19-33}

&&&
\multicolumn{6}{c}{\textbf{Full-caption retrieval}}
& \multicolumn{9}{c|}{\textbf{Reduced-query T$\rightarrow$I retrieval}}
& \multicolumn{6}{c}{\textbf{Full-caption retrieval}}
& \multicolumn{9}{c}{\textbf{Reduced-query T$\rightarrow$I retrieval}} \\

\cmidrule(lr){4-9}
\cmidrule(lr){10-18}
\cmidrule(lr){19-24}
\cmidrule(lr){25-33}

&&&
\multicolumn{3}{c}{\cellcolor{lightpink}{T$\rightarrow$I}}
& \multicolumn{3}{c}{\cellcolor{lightblue}{I$\rightarrow$T}}
& \multicolumn{3}{c}{\cellcolor{salientonly}{Salient-only}}
& \multicolumn{3}{c}{\cellcolor{salienthalf}{Salient$_{50\%}$+Ctx}}
& \multicolumn{3}{c|}{\cellcolor{salientzero}{Context-only}}
& \multicolumn{3}{c}{\cellcolor{lightpink}{T$\rightarrow$I}}
& \multicolumn{3}{c}{\cellcolor{lightblue}{I$\rightarrow$T}}
& \multicolumn{3}{c}{\cellcolor{salientonly}{Salient-only}}
& \multicolumn{3}{c}{\cellcolor{salienthalf}{Salient$_{50\%}$+Ctx}}
& \multicolumn{3}{c}{\cellcolor{salientzero}{Context-only}} \\

&&&
\cellcolor{lightpink}{R@1}
& \cellcolor{lightpink}{R@5}
& \cellcolor{lightpink}{R@10}
& \cellcolor{lightblue}{R@1}
& \cellcolor{lightblue}{R@5}
& \cellcolor{lightblue}{R@10}
& \cellcolor{salientonly}{R@1}
& \cellcolor{salientonly}{R@5}
& \cellcolor{salientonly}{R@10}
& \cellcolor{salienthalf}{R@1}
& \cellcolor{salienthalf}{R@5}
& \cellcolor{salienthalf}{R@10}
& \cellcolor{salientzero}{R@1}
& \cellcolor{salientzero}{R@5}
& \multicolumn{1}{c|}{\cellcolor{salientzero}{R@10}}
& \cellcolor{lightpink}{R@1}
& \cellcolor{lightpink}{R@5}
& \cellcolor{lightpink}{R@10}
& \cellcolor{lightblue}{R@1}
& \cellcolor{lightblue}{R@5}
& \cellcolor{lightblue}{R@10}
& \cellcolor{salientonly}{R@1}
& \cellcolor{salientonly}{R@5}
& \cellcolor{salientonly}{R@10}
& \cellcolor{salienthalf}{R@1}
& \cellcolor{salienthalf}{R@5}
& \cellcolor{salienthalf}{R@10}
& \cellcolor{salientzero}{R@1}
& \cellcolor{salientzero}{R@5}
& \cellcolor{salientzero}{R@10} \\

\midrule

\xmark & \xmark & \xmark
& 8.81 & 24.53 & 35.41
& 7.43 & 24.35 & 36.53
& 7.95 & 22.02 & 32.12
& 7.63 & 22.84 & 33.58
& 3.22 & 9.34 & 15.90
& 81.18 & 96.67 & 98.51
& 80.00 & 96.31 & 98.67
& 64.70 & 86.71 & 91.95
& 73.30 & 92.91 & 96.27
& 29.04 & 50.67 & 58.99 \\

\cmark & \xmark & \xmark
& 9.67 & 25.39 & 38.17
& 9.41 & 25.47 & 36.36
& 8.38 & 23.52 & 34.16
& 9.13 & 25.28 & 35.59
& 4.62 & 10.63 & 16.65
& 82.27 & 97.20 & 98.82
& 80.39 & 96.51 & 98.67
& 68.20 & 89.42 & \textbf{93.75}
& 74.52 & 93.68 & 96.78
& 32.79 & 54.38 & 62.63 \\

\xmark & \cmark & \xmark
& 9.41 & 26.34 & 37.05
& 8.38 & 27.20 & 37.65
& 8.49 & 22.56 & 32.12
& \textbf{9.77} & 25.67 & 35.30
& \textbf{5.26} & 14.18 & 21.59
& 82.88 & 97.18 & 98.76
& 81.22 & 96.90 & 98.59
& 63.44 & 85.95 & 91.81
& \textbf{77.22} & 95.01 & 97.60
& \textbf{43.62} & 67.22 & \textbf{75.37} \\

\cmark & \cmark & \xmark
& 9.76 & 26.60 & 38.26
& 9.50 & 26.77 & 39.64
& \textbf{9.02} & 23.74 & 36.41
& 9.02 & 26.49 & 38.20
& 3.54 & \textbf{15.04} & \textbf{23.09}
& 83.27 & 97.43 & 98.96
& 81.41 & 96.94 & 98.90
& 64.87 & 87.29 & 92.78
& 77.17 & \textbf{95.25} & \textbf{97.69}
& 43.26 & \textbf{67.48} & 75.10 \\

\cmark & \xmark & \cmark
& 9.33 & 25.47 & 37.13
& 7.94 & 25.99 & 37.74
& 8.38 & 24.27 & 33.30
& 8.16 & 24.42 & 35.55
& 2.90 & 10.63 & 17.72
& 81.73 & 96.84 & 98.82
& 79.35 & 96.49 & 98.47
& \textbf{69.21} & \textbf{89.58} & 93.74
& 72.99 & 92.62 & 96.10
& 28.98 & 49.78 & 58.26 \\

\cmark & \cmark & \cmark
& \textbf{10.36} & \textbf{28.32} & \textbf{40.67}
& \textbf{9.84} & \textbf{29.36} & \textbf{40.41}
& 8.27 & \textbf{25.99} & \textbf{37.59}
& 9.06 & \textbf{27.46} & \textbf{38.99}
& 4.51 & 14.93 & 22.13
& \textbf{84.35} & \textbf{97.51} & \textbf{99.18}
& \textbf{82.18} & \textbf{97.14} & \textbf{98.94}
& 68.03 & 88.83 & 93.51
& 74.44 & 93.82 & 97.01
& 42.00 & 66.07 & 73.65 \\

\bottomrule
\end{tabular}
}
\vspace{-20pt}
\end{table*}

\textbf{Saliency-aware alignment losses.}
We ablate the saliency-focused contrastive loss $\salignloss$, context-preserving contrastive loss $\contextloss$, and visual anchoring loss $\anchorloss$. As shown in Tab.~\ref{tab:ablation_loss}, each of the losses boosts retrieval performance. Combining $\salignloss$ and $\contextloss$ further improves most metrics when compared to their individual use, showing the benefit of joint learning from decomposed image-text pairs into salient and non-salient components. Finally, adding $\anchorloss$ on top of $\salignloss + \contextloss$ achieves the best full-caption retrieval performance, especially on INSECT, where T$\rightarrow$I R@10 improves by +3.54\% and I$\rightarrow$T R@5 improves by +3.37\%. 
From the reduced-query results, we can further appreciate the losses that contribute more to the robustness against salient text reduction. Particularly, $\contextloss$ is helpful when salient text is reduced or completely removed (\ie Context-only) by modeling non-salient contextual semantics. Interestingly, simply by introducing $\salignloss$, we could already observe mild improvement in reduced-query settings. These results show that decomposed salient and context views are effective auxiliary supervision, than relying on a single global image-text pair, with $\anchorloss$ further preserving coherence across decomposed views and original images.

\FloatBarrier
\begin{wraptable}{r}{0.4\linewidth}
\vspace{-10pt}
\centering
\caption{\small
\textbf{Ablation on visual anchoring} on
\inlineColorbox{lightpink}{\emph{T$\rightarrow$I}} and
\inlineColorbox{lightblue}{\emph{I$\rightarrow$T}} retrieval,
with correct pairing fixed; highlighted \textbf{best} and
\underline{second best}.
}
\label{tab:anchoring_ablation}
\vspace{-5pt}

\footnotesize
\resizebox{\linewidth}{!}{
\begin{tabular}{@{}cc *{3}{c!{\color{lightlightgray}\vrule width 0.5pt}}c
                    |*{3}{c!{\color{lightlightgray}\vrule width 0.5pt}}c @{}}
\toprule

\multirow{2}{*}{\textbf{Salient}} &
\multirow{2}{*}{\textbf{Context}} &
\multicolumn{4}{c|}{\cellcolor{gray!15}\textbf{INSECT}} &
\multicolumn{4}{c}{\cellcolor{gray!15}\textbf{DOCCI}} \\

\cmidrule(lr){3-6}
\cmidrule(lr){7-10}

&&
\multicolumn{1}{c}{\cellcolor{lightpink}{R@1}} &
\multicolumn{1}{c}{\cellcolor{lightpink}{R@10}} &
\multicolumn{1}{c}{\cellcolor{lightblue}{R@1}} &
\multicolumn{1}{c|}{\cellcolor{lightblue}{R@10}} &
\multicolumn{1}{c}{\cellcolor{lightpink}{R@1}} &
\multicolumn{1}{c}{\cellcolor{lightpink}{R@10}} &
\multicolumn{1}{c}{\cellcolor{lightblue}{R@1}} &
\multicolumn{1}{c}{\cellcolor{lightblue}{R@10}} \\

\midrule

\cmark & \xmark
& \textbf{10.36}
& \textbf{40.67}
& \textbf{9.84}
& \textbf{40.41}
& \textbf{84.35}
& \textbf{99.18}
& \textbf{82.18}
& \textbf{98.94} \\

\xmark & \cmark
& \underline{9.59}
& \underline{39.12}
& 8.55
& \underline{38.60}
& 83.14
& 98.90
& 80.24
& 98.75 \\

\cmark & \cmark
& 9.33
& 37.82
& \underline{9.07}
& 38.43
& \underline{84.18}
& \underline{99.04}
& \underline{81.73}
& \underline{98.86} \\

\bottomrule
\end{tabular}
}
\vspace{-10pt}
\end{wraptable}

\rebuttal{
\textbf{Visual anchoring.}
We compare the default salient-only anchoring with two variants while keeping all other settings fixed: \emph{Context}, which applies visual anchoring between the full image and the contextual view only, and \emph{Salient+Context}, which anchors the full image to both salient and contextual views. Tab.~\ref{tab:anchoring_ablation} shows that salient-only anchoring in \ourmethod achieves the best results across both datasets and retrieval directions. This may be because the salient view reserves a coherent representation of the primary visual content and therefore provides a stable reference for the full image. In contrast, contextual views intentionally represent spatially distributed complementary evidence; forcing them to approximate the full-image representation may weaken that role, while anchoring both views introduces competing consistency constraints.
}

\FloatBarrier
\Needspace{22\baselineskip}

\setlength{\intextsep}{6pt}
\setlength{\columnsep}{12pt}

\begin{wraptable}[13]{r}{0.4\linewidth}
\vspace{-4pt}
\centering
\caption{\small
\textbf{Ablation on decomposition methods} on \inlineColorbox{lightpink}{\emph{T$\rightarrow$I}} and 
\inlineColorbox{lightblue}{\emph{I$\rightarrow$T}}  retrieval, with highlighted \textbf{best} and \underline{second best}.
}
\label{tab:ablation_vt}
\vspace{-5pt}

{\tiny
\setlength{\tabcolsep}{1.0pt}
\renewcommand{\arraystretch}{0.86}
\setlength{\heavyrulewidth}{0.25pt}
\setlength{\lightrulewidth}{0.20pt}
\setlength{\cmidrulewidth}{0.20pt}
\tiny
\begin{tabular}{@{}lcccccccc@{}}
\toprule
\multirow{2}{*}{Method} &
\multicolumn{4}{c}{\cellcolor{gray!15}\textbf{INSECT}} &
\multicolumn{4}{c}{\cellcolor{gray!15}\textbf{DOCCI}} \\
\cmidrule(lr){2-5}
\cmidrule(lr){6-9}
&
\cellcolor{lightpink}{R@1} &
\cellcolor{lightpink}{R@10} &
\cellcolor{lightblue}{R@1} &
\cellcolor{lightblue}{R@10} &
\cellcolor{lightpink}{R@1} &
\cellcolor{lightpink}{R@10} &
\cellcolor{lightblue}{R@1} &
\cellcolor{lightblue}{R@10} \\
\midrule

\rowcolor{gray!8}
\multicolumn{9}{@{}l}{\textit{Text-side variants}} \\
Original text
& \underline{10.36} & 38.08
& 8.64 & 39.38
& 83.02 & 98.94
& 80.35 & 98.67 \\
Random text
& 9.41 & 37.39
& 8.55 & 37.48
& 82.67 & 98.88
& 80.43 & 98.57 \\

\midrule

\rowcolor{gray!8}
\multicolumn{9}{@{}l}{\textit{Visual-side variants}} \\
Random crop
& 9.84 & 37.22
& 8.89 & 37.31
& 83.06 & 98.96
& 80.24 & 98.69 \\
SAM3
& \textbf{10.45} & 38.00
& 8.20 & 38.26
& 82.49 & 99.08
& 81.35 & \underline{98.86} \\
ADMNet
& 8.38 & \underline{39.90}
& 8.20 & 39.38
& 81.88 & 99.00
& 81.14 & 98.82 \\
Spec. Res.
& 9.84 & 39.81
& \underline{9.15} & \underline{39.46}
& 82.61 & 99.08
& 81.22 & 98.82 \\
DeepGaze
& \underline{10.36} & 36.44
& 9.07 & 37.56
& 82.51 & \underline{99.14}
& \underline{81.45} & 98.80 \\

\midrule
\textbf{Ours}
& \underline{10.36} & \textbf{40.67}
& \textbf{9.84} & \textbf{40.41}
& \textbf{84.35} & \textbf{99.18}
& \textbf{82.18} & \textbf{98.94} \\

\bottomrule
\end{tabular}
}

\vspace{4pt}
\end{wraptable}
\textbf{Effectiveness of decomposition methods.}
\rebuttal{All variants use the same backbone and training settings differing only in view construction.}
As shown in Tab.~\ref{tab:ablation_vt}, we analyze different decomposition methods from both the visual and textual sides. 
When fixing salient image (top part of Tab.~\ref{tab:ablation_vt}), substituting salient text with the original caption or a random text weakens overall performance.
When fixing salient text (bottom part of Tab.~\ref{tab:ablation_vt}), replacing salient visual regions with random crops or SAM3 segments~\citep{carion2025sam}. Although SAM3 slightly improves INSECT T$\rightarrow$I R@1 due to its object-like segmentation of compact insect bodies, the gain does not persist at larger recall levels T$\rightarrow$I or in I$\rightarrow$T retrieval. 
Moreover, we also compare our object-level saliency method, with other saliency detection alternatives including ADMNet~\citep{zhou2024admnet} for another object-level saliency, Spectral Residual~\citep{hou2007spectral} for bottom-up frequency-domain saliency, and DeepGaze MSDB~\citep{kummerer2025msdb} for human fixation prediction. Our adopted object-level saliency method (\ie FASA~\citep{yildirim2014fasa}) yields best performance with consistent gains.

\FloatBarrier
\begin{wraptable}{r}{0.6\linewidth}
\vspace{-10pt}
\centering
\caption{\small \textbf{Ablation on saliency-guided decomposition strategies} on \inlineColorbox{lightpink}{\emph{T$\rightarrow$I}} and 
\inlineColorbox{lightblue}{\emph{I$\rightarrow$T}} retrieval, with highlighted \textbf{best} and \underline{second best}.}
\vspace{-5pt}
\label{tab:ablation_modality_independent}
\footnotesize
\resizebox{\linewidth}{!}{
\begin{tabular}{@{}l *{5}{c!{\color{lightlightgray}\vrule width 0.5pt}}c|*{5}{c!{\color{lightlightgray}\vrule width 0.5pt}}c @{}}
\toprule
\multirow{2}{*}{\textbf{Methods}} &
\multicolumn{6}{c|}{\cellcolor{gray!15}\textbf{INSECT}} &
\multicolumn{6}{c}{\cellcolor{gray!15}\textbf{DOCCI}} \\
\cmidrule(lr){2-7}\cmidrule(lr){8-13}
& \multicolumn{1}{c}{\cellcolor{lightpink}{R@1}} & \multicolumn{1}{c}{\cellcolor{lightpink}{R@5}} & \multicolumn{1}{c}{\cellcolor{lightpink}{R@10}}
& \multicolumn{1}{c}{\cellcolor{lightblue}{R@1}} & \multicolumn{1}{c}{\cellcolor{lightblue}{R@5}} & \multicolumn{1}{c|}{\cellcolor{lightblue}{R@10}}
& \multicolumn{1}{c}{\cellcolor{lightpink}{R@1}} & \multicolumn{1}{c}{\cellcolor{lightpink}{R@5}} & \multicolumn{1}{c}{\cellcolor{lightpink}{R@10}}
& \multicolumn{1}{c}{\cellcolor{lightblue}{R@1}} & \multicolumn{1}{c}{\cellcolor{lightblue}{R@5}} & \multicolumn{1}{c}{\cellcolor{lightblue}{R@10}} \\
\midrule
Baseline & 8.81 
 & 24.53 & 35.41  &7.43 & 24.35 & 36.53 & 81.18  &96.67   & 98.51 & 80.00 & 96.31 &98.67 \\
Qwen2.5vl        & \textbf{10.62} & \underline{27.72} & 37.48 & \underline{9.07} & \underline{27.89}  & 38.17 & \underline{83.29}  & \textbf{97.53} & \underline{98.92} & \underline{81.65} & \underline{97.10}  & \underline{98.73} \\
Unipixel           & \underline{10.36} & 26.17 & \underline{37.65} & 8.55 & 27.55 & \underline{38.77}  & 81.59 & 97.16 & 98.82 & 79.08 & 96.47 & 98.35 \\
\textbf{Ours} & \underline{10.36}
 & \textbf{28.32} & \textbf{40.67} &\textbf{9.84} & \textbf{29.36} & \textbf{40.41} & \textbf{84.35}
 & \underline{97.51} & \textbf{99.18} &\textbf{82.18} & \textbf{97.14} & \textbf{98.94} \\
\bottomrule
\end{tabular}
}
\vspace{-10pt}
\end{wraptable}
\textbf{Impacts of saliency-guided decomposition strategies.}
Tab.~\ref{tab:ablation_modality_independent} compares our modality-independent strategy with two cross-modal alternatives. Qwen2.5-VL~\citep{Bai2025Qwen25VLTR} generates detailed text for salient image, while UniPixel~\citep{liu2025unipixel} localizes visual regions from the given salient text. Our modality-independent strategy most often scores the best overall performance, with Qwen2.5-VL being the second best strategy. This result is as expected, as such cross-modal strategy introduces early coupling before training, which might lead to shortcut in learning. Instead, our modality-independent strategy avoids such shortcut in cross-modal alignment, whose decomposed multimodal content are also generally aligned semantically, as analyzed in \supp~Sec.\ref{app:cross_modal_mismatch}.

\section{Conclusion}
\label{sec:conclusion}
In this paper, we characterized saliency collapse, a critical vulnerability in current long-text alignment models as they over-rely on visually prominent descriptions, underutilizing the informative non-salient context. To mitigate saliency collapse, we proposed \ourmethod to decompose image-text pairs into complementary views to form auxiliary objectives that jointly learn both salient and contextual evidence. Extensive evaluations demonstrated that \ourmethod consistently advances long-text retrieval in a plug-and-play manner and exhibits superior robustness under reduced salient information.

\noindent\textbf{Limitations:} While \ourmethod mitigates saliency collapse, it remains an initial exploration of human-centric vision-language alignment. 
Our focus on saliency reduction may not yet encompass the full diversity of human attentional mechanisms. Furthermore, the offline LLM-based text decomposition introduces preprocessing overhead, posing potential scalability challenges for massive datasets.



\section*{Acknowledgments}

This work was supported by Ministero delle Imprese e del Made in Italy (IPCEI Cloud DM 27 giugno 2022 – IPCEI-CL-0000007) and European Union (Next Generation EU). 
We acknowledge the CINECA award under the ISCRA initiative for the availability of high-performance computing resources and support. The authors declare no competing interests.

\newpage
\bibliographystyle{plainnat}
\bibliography{ref} 
\medskip


\newpage
\appendix

\newpage
\section*{Supplementary Material}

This supplementary material provides additional details and analyses to support the main paper. It is organized as follows.

\textbf{Sec.~\ref{sec:dataset_details} }provides detailed descriptions and statistics of the datasets used in our experiments. 

\textbf{Sec.~\ref{sec:generation_validation}} \rebuttal{describes the generation and validation of salient/contextual image--text views, including textual and visual decomposition, output validation, human validation, textual stability, and cross-modal correspondence analyses.}

\rebuttal{
\textbf{Sec.~\ref{sec:controlled_saliency}} presents controlled analyses of saliency collapse, including single-factor controlled deletion and entity-controlled reconstruction.
}

\textbf{Sec.~\ref{sec:query_robustness}} presents additional robustness analyses. 

\textbf{Sec.~\ref{sec:addition_exp}} presents Additional exploratory analyses.

\textbf{Sec.~\ref{sec:qualitative}} provides qualitative comparisons between \ourmethod and best competitor.


\section{Dataset details}
\label{sec:dataset_details}
We conduct experiments on four fine-grained image-text retrieval benchmarks, covering general scenes, fashion and insects. We follow the data splits and preprocessing protocol used in StructXLIP~\citep{ruan2026StructXLIP}. 


DOCCI~\citep{docci2024} is a general-domain image-text dataset containing 15K images. Each image typically contains complex visual attributes and is paired with a human-annotated long English description. The captions average 136 words and are designed to distinguish visually related or similar images. DCI~\citep{urbanek2024picture} is another general-domain long-text image-text dataset, consisting of 7.8K natural images annotated with dense, mask-aligned descriptions averaging over 1,000 words per image. Sketchy~\citep{girella2025lots} is a multimodal fashion dataset extended from Fashionpedia, containing 47K images and 79K garment-level annotations. Each annotation includes a localized sketch, hierarchical attributes, and a garment-level textual description, with an average length of 16 words. INSECT~\citep{truong2025insect} is derived from Insect-1M, a large-scale multimodal insect dataset with over one million insect images covering 34,212 species, together with dense taxonomic labels and insect descriptions. In our experiments, we use a 6K fine-grained biology retrieval split, where each sample is paired with an expert-verified morphological description averaging 80.78 words. 

\begin{figure}[h]
    \centering
    \includegraphics[width=1.0\linewidth]{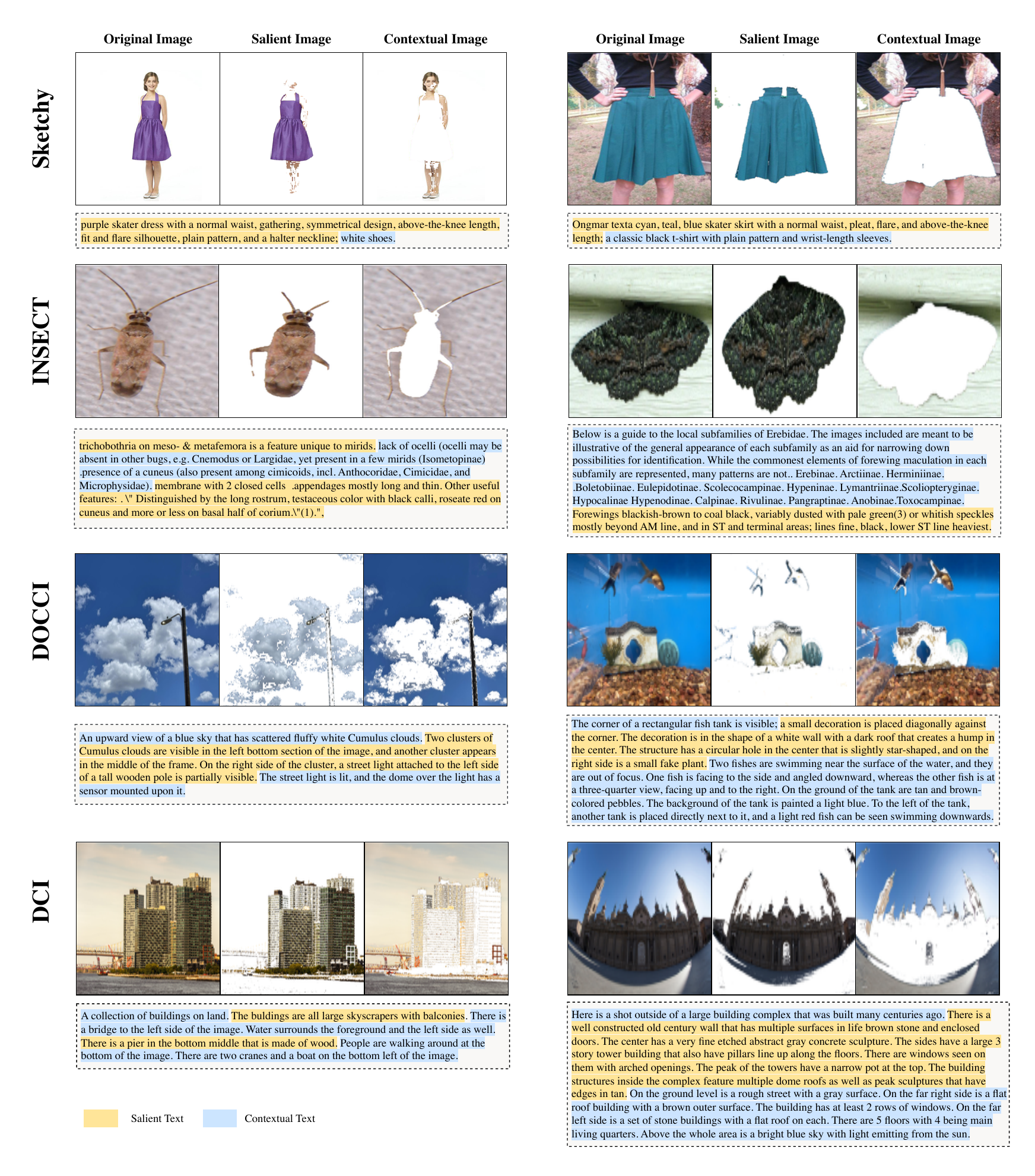}
    \caption{
\textbf{Illustration of decomposed image--text pairs across four datasets.} For each sample from Sketchy, INSECT, DOCCI, and DCI, we show the original image, the salient image, and the contextual image, together with the caption decomposition. Salient text is highlighted in yellow, while contextual text is highlighted in blue.
}
    \label{fig:all_images}
\end{figure}

\begin{figure}[htbp] 
  \centering
  \includegraphics[width=\linewidth]{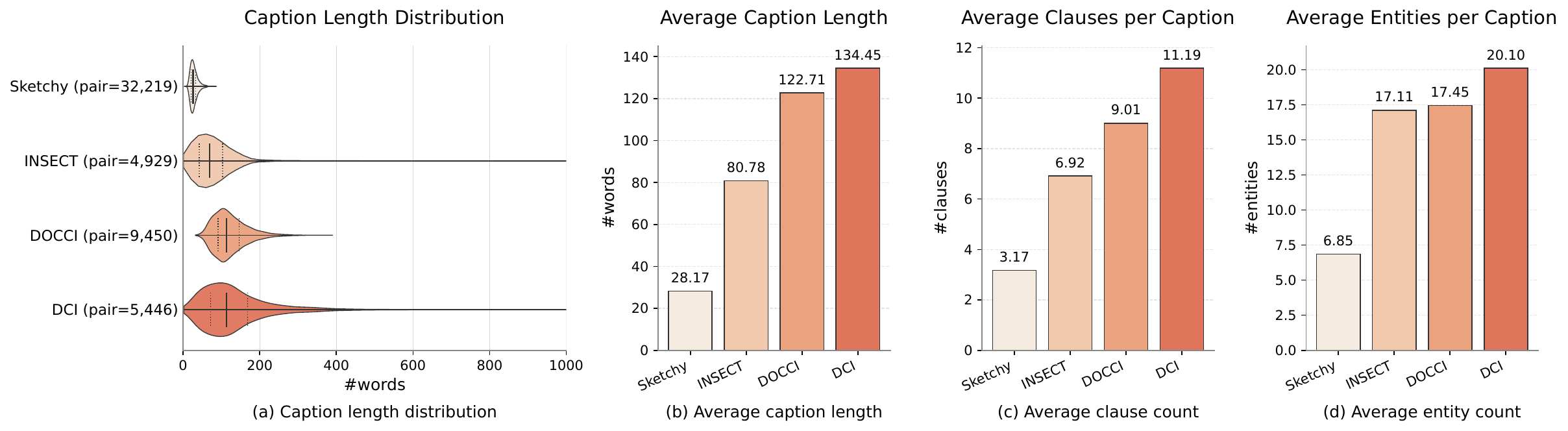} 
  \caption{\textbf{Statistical analysis of the four datasets.} We analyze the textual properties of DOCCI, INSECT, Sketchy, and DCI.
(a) Caption length distribution, where the number in parentheses denotes the number of image-text pairs in each dataset.
(b) Average caption length measured by the number of words.
(c) Average number of clauses per caption.
(d) Average number of semantic entities per caption.}
  \label{fig:dataset_analysis}
\end{figure}

We further provide qualitative examples from the four datasets in Fig.~\ref{fig:all_images}. For each sample, we show the original image, the salient image, the contextual image, and the corresponding salient and contextual textual components. These examples illustrate how the original image-caption pair is decomposed into salient and contextual visual-textual components. In Fig.~\ref{fig:dataset_analysis}, we report dataset statistics, including the number of images, average caption length, average number of clauses per caption, and average number of semantic entities per caption. These statistics provide a more detailed view of the textual density and fine-grained nature of each benchmark.

We define a \textit{semantic entity} as a noun-phrase-level visual concept mentioned in the caption. For example, given the sentence: “A white engine cover and the right wing are visible in the right side of the frame. The wing is straight with a wingtip at the end. To the left is a giant cumulus cloud,” the extracted entities include “engine cover,” “right wing,” “wingtip,” and “cumulus cloud.” We use spaCy, a Python NLP library for automatic text structure analysis, to extract noun phrases with doc.noun\_chunks. We then remove overly generic or non-specific terms to obtain entity statistics that better reflect visually grounded semantic content. As shown in Fig.~\ref{fig:dataset_analysis}(d), all four datasets contain rich semantic entities, making them suitable for evaluating fine-grained image-text alignment.

\section{Salient/Contextual View Generation and Validation}
\label{sec:generation_validation}
In this section, we first describe the textual and visual generation procedures, and then evaluate the resulting decomposition from three aspects: output validity, agreement with human judgments and robustness across textual decomposers, and the semantic correspondence between independently generated visual and textual views.

\subsection{Salient/contextual pair generation}

We independently construct salient and contextual components on the textual and visual sides. This process is performed offline before training and kept fixed across all experiments, so as to avoid introducing additional variation from repeatedly generating texts or regions under different settings. 

\textbf{Textual decomposition.}
On the textual side, we use Qwen2.5~\citep{qwen2025qwen25technicalreport} to perform semantic-level decomposition of the original caption. Given a full caption, the LLM is prompted to divide the description into two parts: salient text and contextual text. Here, salient text refers to descriptions that are directly related to the main visual subject, key local structures, or discriminative attributes in the image. In contrast, contextual text mainly preserves background information, scene context, auxiliary objects, or descriptions with relatively weaker discriminative cues. It is worth noting that contextual text does not indicate useless or incorrect information; rather, it represents complementary semantics relative to the primary matching signal.

\begin{figure}[htbp] 
  \centering
  \includegraphics[width=\linewidth]{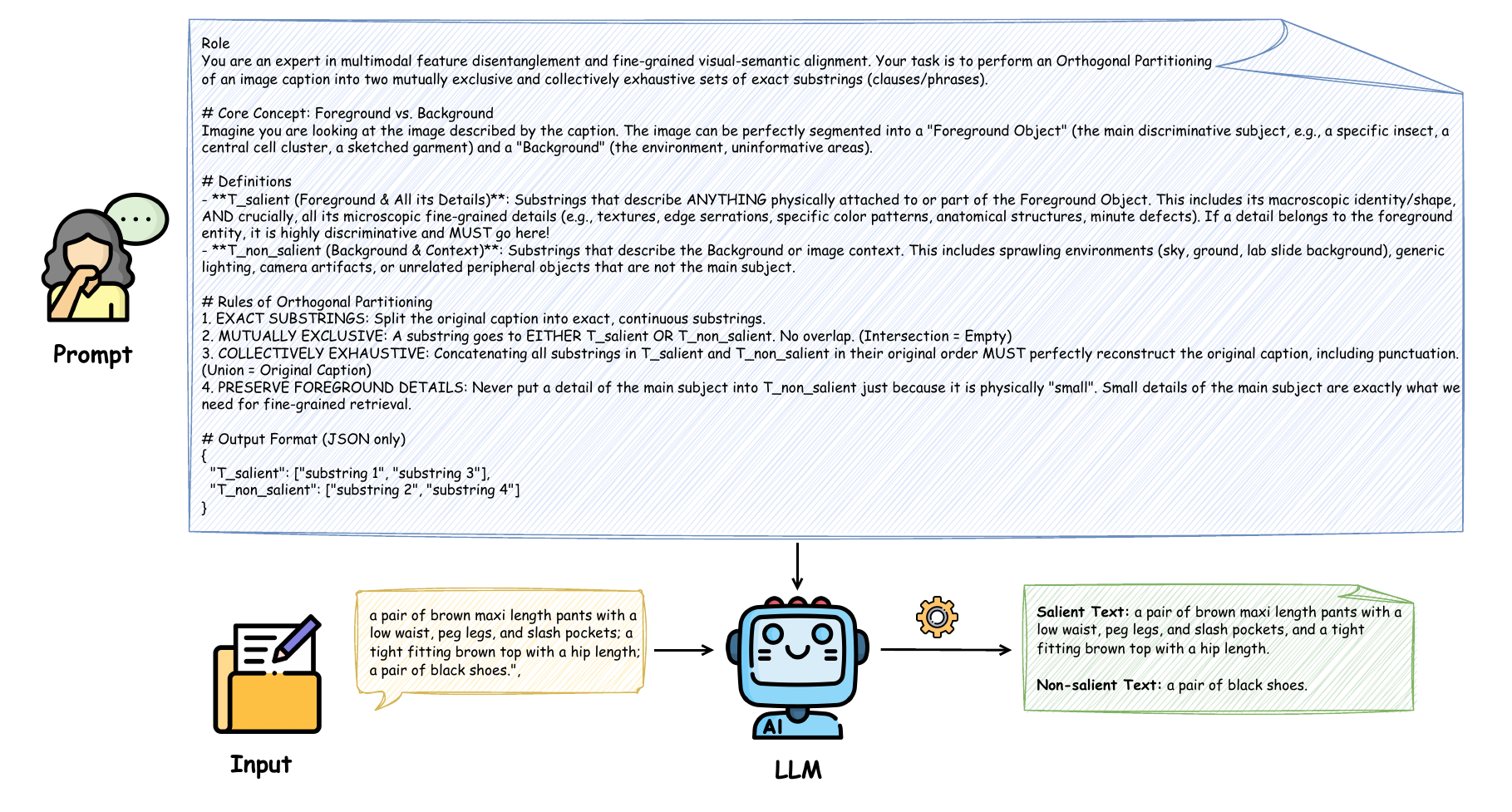} 
  \caption{Prompt and input-output examples. The salient and contextual texts are generated by an LLM during an offline preprocessing stage and kept fixed across all experiments to ensure fair comparisons.
}
  \label{fig:prompt}
\end{figure}

Specifically, we feed the original caption into Qwen2.5 and use a fixed prompt to constrain the output format, asking the model to return salient and contextual textual components. All experiments use the same offline generated decomposed texts to ensure fair comparison across different settings. Fig.~\ref{fig:prompt} shows the prompt used for textual decomposition and an example of the corresponding input and output.
Besides, the time it takes for decomposition of captions into salient and contextual text depends on the dataset size and caption length. For example, for DOCCI with 9450 images, it is about 15 hrs, while for INSECT with 4929 images, it is 7 hrs.

\textbf{Visual decomposition.}
On the visual side, we use an object saliency detection method FASA~\citep{yildirim2014fasa} to extract salient regions from the original image. For each image, we first generate a saliency map and convert it into a salient-region mask. Based on this mask, the salient image is constructed by preserving the salient regions in the original image, while the contextual image is obtained from the complementary regions after removing the salient part. In this way, the original image is decomposed into two visual views: the salient image emphasizes the primary visual evidence, whereas the contextual image retains contextual visual information. Together with the textual decomposition, this provides a more structured image-text supervision signal for training.
For visual decomposition, FASA processes each image in about 2 seconds on average. In comparison, SAM3-based~\citep{carion2025sam} region generation is substantially more expensive, requiring about 6 seconds per image on average. Importantly, these costs occur only once before fine-tuning, and \ourmethod does not introduce any additional computational overhead during inference.

\rebuttal{
\textbf{Output validation.}
The LLM is constrained to preserve the original caption content and return a fixed JSON decomposition, with free generation or rewriting disallowed. Each output is then validated before training for three failure modes: (i) invalid JSON format, (ii) content unsupported by the source caption, and (iii) incomplete reconstruction of the original caption. Across the four datasets, we observe no JSON-format failures. Before regeneration, the generated partitions achieve 98.04\% mean per-sample caption coverage and 85.85\% exact complete coverage. Failed cases are automatically detected and regenerated before training.
}

\rebuttal{\subsection{Human Validation}}
\label{sec:human_validation}

\textbf{Human visual-grounding validation.}
Three annotators evaluate 150 samples, with 50 images each from DOCCI, DCI, and Sketchy. For each textual component, they judge (i) \emph{visual verifiability}: whether the description is observable in the corresponding image, and (ii) \emph{visual discriminability}: whether the description identifies the ground-truth image against the highest-ranked incorrect retrieval.
As shown in Tab.~\ref{tab:human_validation_sc}, 
human annotators judge 84.00\% of salient descriptions and 84.67\% of contextual descriptions as visually observable. In the two-image discrimination task, salient and contextual text identify the ground-truth image in 94.00\% and 90.67\% of cases, with 83.11\% and 78.89\% annotator agreement, respectively. These results independently verify that the LLM-generated salient text corresponds to visually grounded and discriminative content, while the contextual text not only retains sample-specific information, but also contains visually observable and discriminative evidence.

\begin{table}[h]
\centering
\small
\caption{\textbf{Human validation of salient and contextual text.}
Results report visual verifiability, visual discriminability, and annotator agreement (\%).}
\label{tab:human_validation_sc}
\setlength{\tabcolsep}{8pt}
\begin{tabular}{lcc}
\toprule
\textbf{Metric} & \textbf{Salient} & \textbf{Contextual} \\
\midrule
Visual verifiability    & 84.00 & 84.67 \\
Visual discriminability & 94.00 & 90.67 \\
Annotator agreement     & 83.11 & 78.89 \\
\bottomrule
\end{tabular}
\end{table}

\textbf{Saliency-span agreement.}
Ten annotators label all 5100 DOCCI test samples, with 510 samples assigned to each annotator, by marking the caption span describing the most visually salient content at first glance. Mapping the human and LLM spans to the original token positions, the LLM split covers 92.74\% of human-salient tokens and achieves 77.58\% token F1 / 69.04\% IoU, substantially outperforming an equal-length random-span baseline (47.88\% / 36.96\%). These results show substantial agreement between the LLM decomposition and independently perceived visual saliency.

\rebuttal{\subsection{Prompt and LLM Stability}}
We evaluate the sensitivity of textual decomposition to the prompt and LLM. Using a substantially rewritten Qwen2.5~\citep{qwen2025qwen25technicalreport} prompt and, separately, Llama-3.1~\citep{grattafiori2025llama} with the original prompt, the CLIP semantic similarity between corresponding outputs ranges from 0.808--0.912 across prompts and 0.841--0.909 across LLMs, indicating substantial semantic stability.

We further regenerate the training partitions with Llama-3.1 and retrain \ourmethod with all other settings fixed. As shown in Tab.~\ref{tab:text_decomposer_ablation}, retrieval performance remains comparable after replacing the default Qwen2.5 decomposer.

\begin{table}[t]
\centering
\caption{\small
\textbf{Robustness to the textual decomposer.}
R@1 and R@10 (\%) on
\inlineColorbox{lightpink}{\emph{T$\rightarrow$I}} and
\inlineColorbox{lightblue}{\emph{I$\rightarrow$T}} retrieval.
\ourmethod{} is retrained using partitions generated by each LLM,
with all other settings fixed.
Best results are \textbf{bold}; second best \underline{underlined}.
$\Delta$ denotes the margin of Qwen2.5 over Llama-3.1.
}
\label{tab:text_decomposer_ablation}
\vspace{2pt}

{\scriptsize
\setlength{\tabcolsep}{3.0pt}
\renewcommand{\arraystretch}{0.95}
\setlength{\heavyrulewidth}{0.25pt}
\setlength{\lightrulewidth}{0.20pt}
\setlength{\cmidrulewidth}{0.20pt}

\begin{tabular}{@{}lcccccccc@{}}
\toprule
\multirow{2}{*}{Text decomposer} &
\multicolumn{4}{c}{\cellcolor{gray!15}\textbf{INSECT}} &
\multicolumn{4}{c}{\cellcolor{gray!15}\textbf{DOCCI}} \\
\cmidrule(lr){2-5}
\cmidrule(lr){6-9}
&
\cellcolor{lightpink}{R@1} &
\cellcolor{lightpink}{R@10} &
\cellcolor{lightblue}{R@1} &
\cellcolor{lightblue}{R@10} &
\cellcolor{lightpink}{R@1} &
\cellcolor{lightpink}{R@10} &
\cellcolor{lightblue}{R@1} &
\cellcolor{lightblue}{R@10} \\
\midrule

Llama-3.1
& \underline{10.28} & \underline{38.08}
& \underline{9.59} & \underline{38.60}
& \underline{83.33} & \underline{98.84}
& \underline{81.90} & \underline{98.73} \\

\midrule

\textbf{Qwen2.5 (default)}
& \textbf{10.36} & \textbf{40.67}
& \textbf{9.84} & \textbf{40.41}
& \textbf{84.35} & \textbf{99.18}
& \textbf{82.18} & \textbf{98.94} \\

\midrule

\textit{$\Delta$}
& \gain{0.08} & \gain{2.59}
& \gain{0.25} & \gain{1.81}
& \gain{1.02} & \gain{0.34}
& \gain{0.28} & \gain{0.21} \\

\bottomrule
\end{tabular}
}
\end{table}

\subsection{Cross-modal Correspondence Analysis}

\rebuttal{\subsubsection{Salient–contextual swapping control}}
\rebuttal{
\textbf{Salient–contextual pairing.}
To evaluate whether the gains depend on the correspondence between auxiliary views, we pair salient images with contextual text and contextual images with salient text, keeping the decomposed views,  training settings, and computation remain fixed. As shown in Tab.~\ref{tab:pairing_ablation}, correct matching wins all eight metrics, including +2.31/+2.81 DOCCI T→I/I→T R@1. Because the variants differ only in correspondence, arbitrary partial-view alignment cannot explain the gains; preserving the intended salient–contextual roles is essential. 
}

\begin{table}[t]
\centering
\caption{\small
\textbf{Ablation on salient--contextual swapping control.}
Results on
\inlineColorbox{lightpink}{\emph{T$\rightarrow$I}} and
\inlineColorbox{lightblue}{\emph{I$\rightarrow$T}} retrieval,
with salient-only anchoring fixed.
Best results are \textbf{bold}.
}
\label{tab:pairing_ablation}

\vspace{2pt}

{\scriptsize
\setlength{\tabcolsep}{1.8pt}
\renewcommand{\arraystretch}{0.95}

\resizebox{0.7\linewidth}{!}{%
\begin{tabular}{@{}lcccccccc@{}}
\toprule

\multirow{2}{*}{\textbf{Pairing}} &
\multicolumn{4}{c}{\cellcolor{gray!15}\textbf{INSECT}} &
\multicolumn{4}{c}{\cellcolor{gray!15}\textbf{DOCCI}} \\

\cmidrule(lr){2-5}
\cmidrule(lr){6-9}

&
\cellcolor{lightpink}{R@1} &
\cellcolor{lightpink}{R@10} &
\cellcolor{lightblue}{R@1} &
\cellcolor{lightblue}{R@10} &
\cellcolor{lightpink}{R@1} &
\cellcolor{lightpink}{R@10} &
\cellcolor{lightblue}{R@1} &
\cellcolor{lightblue}{R@10} \\

\midrule

Swapped
& 9.76
& 38.43
& 9.67
& 39.46
& 82.04
& 98.96
& 79.37
& 98.69 \\

\textbf{Correctly matched}
& \textbf{10.36}
& \textbf{40.67}
& \textbf{9.84}
& \textbf{40.41}
& \textbf{84.35}
& \textbf{99.18}
& \textbf{82.18}
& \textbf{98.94} \\

\midrule

\textit{$\Delta$}
& \gain{0.60}
& \gain{2.24}
& \gain{0.17}
& \gain{0.95}
& \gain{2.31}
& \gain{0.22}
& \gain{2.81}
& \gain{0.25} \\

\bottomrule
\end{tabular}%
}
}
\end{table}

\subsubsection{Robustness to cross-modal mismatch}
\label{app:cross_modal_mismatch}
In our method, we generate the salient/contextual image and text independently. This raises a natural concern: the resulting salient/contextual image-text pairs may be semantically mismatched and may introduce noisy auxiliary supervision during fine-tuning. For example, if a salient text does not actually correspond to the salient image region, fine-tuning would incorrectly pull an unmatched image-text component pair closer, potentially harming the learning.

To examine this risk, we use a frozen CLIP~\citep{radford2021learning} model to compute the initial agreement score for each salient/contextual image-text pair. We then independently filter the lowest-scoring 5\%, 10\%, and 20\% pairs in the salient and contextual branches. The original image-text pairs are always retained; filtering only affects the corresponding auxiliary losses. 

As shown in Table~\ref{tab:agreement}, the full model without filtering achieves the best overall performance on DOCCI, INSECT. 
These results suggest that the independently generated salient/contextual pairs are not dominated by semantic mismatches. 
The fact that filtering often degrades performance, indicates that our modality-independent strategy can generally produce pairs that are useful for long-text alignment.

Recent work~\citep{han2025learning} shows LLMs can acquire visual semantic priors from text-only pre-training and that such priors can be decomposed into perception and reasoning priors with different origins. In particular, perception-oriented ability depends more on the vision encoder and visual instruction tuning while visual reasoning is strongly shaped by reasoning priors acquired during language pre-training. It motivates why salient text extraction from captions can be viewed as a language-side visual-semantic reasoning process, rather than simple string splitting. Therefore, the independently generated salient/contextual images and texts show semantic consistency in terms of visual perception.


\begin{table}[htbp]
\centering
\caption{\textbf{Robustness to cross-modal mismatch.} Results on \inlineColorbox{lightpink}{\emph{Text$\rightarrow$Image}} and 
\inlineColorbox{lightblue}{\emph{Image$\rightarrow$Text}} retrieval for INSECT, DOCCI. We discard different proportions of generated pairs according to their similarity scores. For example, 5\% filtering ratio means removing the lowest-scored 5\% pairs.}
\label{tab:agreement}

\resizebox{\columnwidth}{!}{
\begin{tabular}{@{}l *{5}{c!{\color{lightlightgray}\vrule width 0.5pt}}c|*{5}{c!{\color{lightlightgray}\vrule width 0.5pt}}c @{}}
\toprule
\multirow{2}{*}{\textbf{Filtering ratio}} &
\multicolumn{6}{c|}{\cellcolor{gray!15}\textbf{INSECT}} &
\multicolumn{6}{c}{\cellcolor{gray!15}\textbf{DOCCI}} \\
\cmidrule(lr){2-7}\cmidrule(lr){8-13}
& \multicolumn{1}{c}{\cellcolor{lightpink}{R@1}} & \multicolumn{1}{c}{\cellcolor{lightpink}{R@5}} & \multicolumn{1}{c}{\cellcolor{lightpink}{R@10}}
& \multicolumn{1}{c}{\cellcolor{lightblue}{R@1}} & \multicolumn{1}{c}{\cellcolor{lightblue}{R@5}} & \multicolumn{1}{c|}{\cellcolor{lightblue}{R@10}}
& \multicolumn{1}{c}{\cellcolor{lightpink}{R@1}} & \multicolumn{1}{c}{\cellcolor{lightpink}{R@5}} & \multicolumn{1}{c}{\cellcolor{lightpink}{R@10}}
& \multicolumn{1}{c}{\cellcolor{lightblue}{R@1}} & \multicolumn{1}{c}{\cellcolor{lightblue}{R@5}} & \multicolumn{1}{c}{\cellcolor{lightblue}{R@10}} \\
\midrule
20\%      &\underline{10.02}	&27.63	&37.39	&9.24	&26.08	&38.51&\underline{80.25}&96.71&98.69&78.37&\underline{96.37}&\underline{98.57}\\
10\%   &\underline{10.02}	&26.34	&37.74&	\underline{9.50}	&26.42	&36.96&79.63 &96.65&98.47&78.35&96.22&98.37\\
5\% & 9.84	&\textbf{29.02}	&\underline{39.72}&	8.98&	\underline{27.89}&	\underline{39.12}&\underline{80.25}&\underline{96.76}&\underline{98.82}&\underline{79.16} &95.86&98.53\\
0\% (Ours) & \textbf{10.36}
 & \underline{28.32} & \textbf{40.67} &\textbf{9.84} & \textbf{29.36} & \textbf{40.41} & \textbf{84.35}
 & \textbf{97.51} & \textbf{99.18} &\textbf{82.18} & \textbf{97.14} & \textbf{98.94} \\
\bottomrule


\end{tabular}
}
\end{table}

\rebuttal{
\section{Controlled Analyses of Saliency Collapse}
\label{sec:controlled_saliency}
}
Beyond the length-matched analysis in Fig.~\ref{fig:lengthcontrol}, we further control for several textual factors that may differ between salient and contextual descriptions, including entity count, lexical statistics, caption position, and text-only discriminability. We study these factors through single-factor controlled deletion and entity-controlled reconstruction under a fixed query template.

\subsection{Single-factor controlled deletion}
\label{sec:controlled_deletion}

For each salient deletion, we enumerate combinations of complete contextual clauses from the same caption and select a contextual deletion matched separately by:
(i) spaCy noun-phrase/entity count;
(ii) TF--IDF density computed from training captions;
(iii) normalized caption position; or
(iv) E5-large-v2 text-only discriminability.
These are four separate controls, rather than a joint match on all four properties.
We evaluate LongCLIP-FT~\citep{zhang2024long}, GOAL~\citep{goal2025}, and StructXLIP~\citep{ruan2026StructXLIP} on DOCCI~\citep{docci2024} and INSECT~\citep{truong2025insect} at salient-text deletion ratios of
$\mathcal{D}=\{25\%,50\%,75\%,90\%,100\%\}$.
We report the average difference between contextual-deletion and salient-deletion text-to-image R@1.
Table~\ref{tab:controlled_deletion} shows positive average gaps for all six model--dataset settings under every matching metric.
Thus, matching any one of the tested factors does not eliminate the observed average asymmetry between salient and contextual deletion.

\begin{table}[t]
\centering
\small
\caption{
\textbf{Single-factor controlled deletion.}
Each entry reports contextual-deletion R@1 - salient-deletion R@1, averaged over five salient-text deletion ratios.
Columns represent separate matching metric.
Positive values indicate greater degradation under salient deletion.
}
\label{tab:controlled_deletion}
\begin{tabular}{llrrrr}
\toprule
Dataset & Model & Entity & TF--IDF & Position & E5 \\
\midrule
DOCCI
& GOAL        & +13.95 & +17.25 & +16.33 & +17.32 \\
& LongCLIP-FT & +14.15 & +17.54 & +16.68 & +17.58 \\
& StructXLIP  & +13.14 & +17.26 & +15.98 & +17.39 \\
\midrule
INSECT
& GOAL        & +2.47 & +2.02 & +3.09 & +2.57 \\
& LongCLIP-FT & +2.41 & +2.51 & +3.47 & +2.59 \\
& StructXLIP  & +3.15 & +2.83 & +3.58 & +3.32 \\
\bottomrule
\end{tabular}
\end{table}

\subsection{Entity-controlled reconstruction}
\label{sec:entity_reconstruction}

We further isolate saliency from query structure using entity-controlled reconstruction. We fix the syntax, entity positions, and entity count with the template
\begin{quote}
\texttt{An image containing: entity 1, ..., entity K}.
\end{quote}
We then progressively replace salient entities with contextual entities from the same caption, matching replacement pairs by CLIP-token length and IDF computed from the training captions. We use $K=4$ and replacement ratios of ${0\%,25\%,50\%,75\%,100\%}$, where $0\%$ contains only salient entities and $100\%$ replaces all four with contextual entities. The resulting query-length difference is only 0.73 tokens, while the entity-level IDF differs by 4.1\% (standardized difference: 0.187).
We evaluate LongCLIP-FT~\citep{zhang2024long}, GOAL~\citep{goal2025}, StructXLIP~\citep{ruan2026StructXLIP}, and \ourmethod on T$\rightarrow$I R@1 over three seeds. As shown in Tab.~\ref{tab:entity_reconstruction}, retrieval performance consistently decreases as salient entities are replaced by matched contextual entities. \ourmethod achieves the highest mean R@1 at every replacement ratio and demonstrates a smaller endpoint drop than the baselines. For example, StructXLIP decreases from 20.14\% to 9.42\% ($-10.72\%$), whereas \ourmethod decreases from 21.02\% to 13.53\% ($-7.49\%$).
These results support RoSA's improved utilization of contextual evidence under controlled query reconstruction.

\begin{table}[t]
\centering
\small
\setlength{\tabcolsep}{4pt}
\caption{
Entity-controlled reconstruction with $K=4$.
Columns indicate the percentage of salient entities replaced by matched contextual entities from the same caption.
Entries report text-to-image R@1 (\%) over three seeds.
The best mean in each column is highlighted in \textbf{bold}.
}
\label{tab:entity_reconstruction}
\begin{tabular}{lccccc}
\toprule
Model & 0\% & 25\% & 50\% & 75\% & 100\% \\
\midrule
LongCLIP-FT
& $18.26 \pm 0.83$
& $15.48 \pm 0.20$
& $13.09 \pm 0.55$
& $10.66 \pm 0.12$
& $7.77 \pm 0.36$ \\
GOAL
& $20.06 \pm 0.87$
& $17.85 \pm 0.55$
& $15.58 \pm 0.49$
& $13.13 \pm 0.13$
& $9.67 \pm 0.46$ \\
StructXLIP
& $20.14 \pm 0.86$
& $17.62 \pm 0.21$
& $15.47 \pm 0.18$
& $12.88 \pm 0.34$
& $9.42 \pm 0.01$ \\
\textbf{RoSA}
& $\mathbf{21.02 \pm 0.31}$
& $\mathbf{19.65 \pm 0.10}$
& $\mathbf{17.39 \pm 0.16}$
& $\mathbf{15.98 \pm 0.27}$
& $\mathbf{13.53 \pm 0.48}$ \\
\bottomrule
\end{tabular}
\end{table}

\textbf{Discussion.}
Across all 39/39 baseline endpoint comparisons, we observe a consistent saliency-reliance gap, while \ourmethod consistently narrows this gap. Together with the deletion controls, these results substantially reduce the influence of textual confounders: the reconstruction fixes the query template, entity count, and entity positions while closely matching token length and IDF, supporting concentrated reliance on salient evidence and \ourmethod's improved use of the remaining context.


\section{Robustness to alternative query construction}
\label{sec:query_robustness}

Training uses decomposed image--text views alongside the original pairs, whereas evaluation always retrieves the original full images without involving decomposed visual views. To test the remaining text-side overlap, we freeze all checkpoints and regenerate only the test queries using: (i) Qwen2.5~\citep{qwen2025qwen25technicalreport} with a substantially rewritten prompt, (ii) an unseen Llama-3.1~\citep{grattafiori2025llama} with the original prompt, and (iii) human construction without an LLM. For the latter, ten annotators partition the complete 5,100-sample DOCCI~\citep{docci2024} test set, with 510 samples assigned to each annotator. The rewritten-prompt and Llama-3.1 settings are evaluated on DOCCI and INSECT~\citep{truong2025insect}, while human queries are evaluated on DOCCI. 

For each construction, we evaluate the same three query conditions as in the main robustness experiment: \emph{Salient-only}, \emph{Salient$_{50\%}$+Context}, and \emph{Context-only}. Tab.~\ref{tab:independent_queries} reports T$\rightarrow$I R@1 and R@10 without retraining either model.
Across the original construction in Tab.~\ref{tab:salient_robustness} and the three alternatives, \ourmethod wins 39/42 comparisons and all 14/14 Context-only comparisons. Its advantage remains largest when salient evidence is absent, including under human queries with no LLM in the pipeline. On DOCCI, the Context-only R@1 gains are 12.18\%, 9.30\%, and 8.84\% for the rewritten prompt, Llama-3.1, and human queries, respectively; on INSECT, the corresponding gains are 0.10\% and 1.98\%. These results separate the robustness gains from adaptation to a particular prompt or decomposition rule.

\begin{table}[t]
\centering
\caption{
\textbf{Robustness to alternative query construction.}
We report \inlineColorbox{lightpink}{\emph{T$\rightarrow$I}}
Recall@K (\%) on INSECT and DOCCI using test queries reconstructed
with a rewritten prompt, Llama-3.1, or human annotations under
\inlineColorbox{salientonly}{\emph{Salient-only}},
\inlineColorbox{salienthalf}{\emph{Salient$_{50\%}$+Context}}, and
\inlineColorbox{salientzero}{\emph{Context-only}}.
Model checkpoints and full-image retrieval galleries remain fixed.
For each query construction, we highlight \textbf{best} and
\underline{second best} results;
$\Delta$ denotes the margin of \ourmethod{} over StructXLIP (\%).
Human queries are evaluated only on DOCCI; ``--'' indicates
an unevaluated setting.
}
\label{tab:independent_queries}

\scriptsize
\setlength{\tabcolsep}{1.5pt}
\renewcommand{\arraystretch}{1.05}

\resizebox{\textwidth}{!}{
\begin{tabular}{@{} l l *{6}{c} | *{6}{c} @{}}
\toprule

\multirow{3}{*}{\textbf{Query construction}}
& \multirow{3}{*}{\textbf{Method}}
& \multicolumn{6}{c|}{\cellcolor{panelgray}\textbf{INSECT}}
& \multicolumn{6}{c}{\cellcolor{panelgray}\textbf{DOCCI}} \\

\cmidrule(lr){3-8}
\cmidrule(lr){9-14}

& &
\multicolumn{2}{c}{\cellcolor{salientonly}{\textbf{Salient-only}}}
& \multicolumn{2}{c}{\cellcolor{salienthalf}{\textbf{Salient$_{50\%}$+Context}}}
& \multicolumn{2}{c|}{\cellcolor{salientzero}{\textbf{Context-only}}}

& \multicolumn{2}{c}{\cellcolor{salientonly}{\textbf{Salient-only}}}
& \multicolumn{2}{c}{\cellcolor{salienthalf}{\textbf{Salient$_{50\%}$+Context}}}
& \multicolumn{2}{c}{\cellcolor{salientzero}{\textbf{Context-only}}} \\

\cmidrule(lr){3-4}
\cmidrule(lr){5-6}
\cmidrule(lr){7-8}
\cmidrule(lr){9-10}
\cmidrule(lr){11-12}
\cmidrule(lr){13-14}

& &
\cellcolor{salientonly}{R@1}
& \cellcolor{salientonly}{R@10}
& \cellcolor{salienthalf}{R@1}
& \cellcolor{salienthalf}{R@10}
& \cellcolor{salientzero}{R@1}
& \multicolumn{1}{c|}{\cellcolor{salientzero}{R@10}}

& \cellcolor{salientonly}{R@1}
& \cellcolor{salientonly}{R@10}
& \cellcolor{salienthalf}{R@1}
& \cellcolor{salienthalf}{R@10}
& \cellcolor{salientzero}{R@1}
& \cellcolor{salientzero}{R@10} \\

\midrule


\multirow{3}{*}{Rewritten prompt}
& StructXLIP {\tiny [CVPR'26]}
& \underline{7.55} & \underline{33.19}
& \textbf{9.68} & \underline{36.17}
& \underline{3.62} & \underline{16.81}

& \underline{65.15} & \underline{92.53}
& \underline{78.30} & \underline{97.47}
& \underline{34.98} & \underline{65.85} \\

& \textbf{\ourmethod{} (Ours)}
& \textbf{7.98} & \textbf{35.11}
& \underline{9.47} & \textbf{39.04}
& \textbf{3.72} & \textbf{22.66}

& \textbf{65.81} & \textbf{93.13}
& \textbf{79.83} & \textbf{98.28}
& \textbf{47.16} & \textbf{78.34} \\

\cmidrule(lr){2-8}
\cmidrule(lr){9-14}

& \textit{$\Delta$}
& \gain{0.43} & \gain{1.92}
& \down{0.21} & \gain{2.87}
& \gain{0.10} & \gain{5.85}

& \gain{0.66} & \gain{0.60}
& \gain{1.53} & \gain{0.81}
& \gain{12.18} & \gain{12.49} \\

\midrule


\multirow{3}{*}{Llama-3.1}
& StructXLIP {\tiny [CVPR'26]}
& \textbf{7.19} & \underline{34.69}
& \underline{9.17} & \underline{35.83}
& \underline{3.12} & \underline{18.02}

& \underline{66.92} & \underline{93.23}
& \underline{76.04} & \underline{96.82}
& \underline{35.29} & \underline{64.47} \\

& \textbf{\ourmethod{} (Ours)}
& \underline{6.67} & \textbf{35.94}
& \textbf{9.90} & \textbf{38.54}
& \textbf{5.10} & \textbf{21.04}

& \textbf{67.86} & \textbf{93.58}
& \textbf{78.25} & \textbf{97.43}
& \textbf{44.59} & \textbf{76.02} \\

\cmidrule(lr){2-8}
\cmidrule(lr){9-14}

& \textit{$\Delta$}
& \down{0.52} & \gain{1.25}
& \gain{0.73} & \gain{2.71}
& \gain{1.98} & \gain{3.02}

& \gain{0.94} & \gain{0.35}
& \gain{2.21} & \gain{0.61}
& \gain{9.30} & \gain{11.55} \\

\midrule


\multirow{3}{*}{Human}
& StructXLIP {\tiny [CVPR'26]}
& -- & --
& -- & --
& -- & --

& \underline{56.44} & \underline{86.58}
& \underline{81.20} & \underline{98.32}
& \underline{50.01} & \underline{78.94} \\

& \textbf{\ourmethod{} (Ours)}
& -- & --
& -- & --
& -- & --

& \textbf{58.55} & \textbf{88.19}
& \textbf{82.05} & \textbf{98.87}
& \textbf{58.85} & \textbf{86.71} \\

\cmidrule(lr){2-8}
\cmidrule(lr){9-14}

& \textit{$\Delta$}
& -- & --
& -- & --
& -- & --

& \gain{2.11} & \gain{1.61}
& \gain{0.85} & \gain{0.55}
& \gain{8.84} & \gain{7.77} \\

\bottomrule
\end{tabular}
}
\vspace{-15pt}
\end{table}

\section{Additional exploratory analyses}
\label{sec:addition_exp}

\subsection{Cross-domain Evaluation on PathGen}
\label{sec:pathgen}

We further evaluate our method on PathGen-1.6M~\citep{sun2024pathgen}, a pathology image-text dataset constructed from approximately 7,800 whole-slide images (WSIs) and containing over 1.6 million patch--caption pairs. 
Despite its large scale, the dataset exhibits substantial redundancy: within the same WSI, many extracted patches are visually similar and their associated captions are often repetitive, focusing on a limited set of pathological patterns. Consequently, the increase in dataset size does not necessarily translate into proportional semantic diversity.

Moreover, in realistic clinical and research settings, obtaining million-scale image--text annotations is impractical. Pathology experts are more likely to provide a limited number of annotations covering diverse pathological themes rather than repeatedly describing the same lesion region. Therefore, for downstream fine-tuning tasks, constructing a compact yet semantically diverse dataset is both practically meaningful and computationally desirable.

Motivated by these observations, we perform semantic condensation on PathGen-1.6M through a two-stage pipeline. 
First, we conduct stratified splitting at the WSI level based on TCGA TSS code mappings. WSIs are divided into training and testing sets with an 80\%/20\% ratio while strictly ensuring that all patch--caption pairs from the same WSI remain within the same split, thereby preventing data leakage.

Second, semantic de-redundancy is performed within each WSI. Specifically, all captions are encoded using S-BioBERT~\citep{deka2021unsupervised}, followed by agglomerative hierarchical clustering with cosine distance to group semantically similar captions into thematic clusters. The clustering threshold is set to 0.4, approximately corresponding to a cosine similarity greater than 0.6. Within each cluster, only the patch--caption pair closest to the cluster centroid is retained as the representative sample, while the remaining samples are discarded as redundant. This process effectively removes repetitive descriptions while preserving the major pathological semantic themes within each WSI.

After semantic condensation, the final dataset contains 15,749 training pairs and 4,037 testing pairs. 
Based on these selected image--caption pairs, we construct a pathology retrieval benchmark in which captions primarily describe cellular structures, tissue morphology, and local pathological patterns, with an average caption length of 51 words.

\begin{table}[t]
\centering
\caption{
\textbf{Cross-modal retrieval on PathGen.}
We report R@K under \inlineColorbox{lightpink}{\emph{T$\rightarrow$I}} and 
\inlineColorbox{lightblue}{\emph{I$\rightarrow$T}} retrieval.
Best and second-best results are marked in \textbf{bold} and \underline{underline}.
}
\label{tab:pathgen}
\scriptsize
\setlength{\tabcolsep}{4pt}
\renewcommand{\arraystretch}{0.9}

\begin{tabular*}{\textwidth}{@{\extracolsep{\fill}} l *{2}{c!{\color{lightlightgray}\vrule width 0.5pt}}c|*{2}{c!{\color{lightlightgray}\vrule width 0.5pt}}c @{}}
\toprule
\multirow{2}{*}{\shortstack[c]{\textbf{Methods}}} &
\multicolumn{3}{c|}{\cellcolor{gray!15}\textbf{T$\rightarrow$I}} &
\multicolumn{3}{c}{\cellcolor{gray!15}\textbf{I$\rightarrow$T}} \\
\cmidrule(lr){2-4}\cmidrule(lr){5-7}
& \multicolumn{1}{c}{\cellcolor{lightpink}{R@1}}
& \multicolumn{1}{c}{\cellcolor{lightpink}{R@5}}
& \multicolumn{1}{c|}{\cellcolor{lightpink}{R@10}}
& \multicolumn{1}{c}{\cellcolor{lightblue}{R@1}}
& \multicolumn{1}{c}{\cellcolor{lightblue}{R@5}}
& \multicolumn{1}{c}{\cellcolor{lightblue}{R@10}} \\
\midrule
Long-CLIP {\tiny [ECCV'24]}
& 3.20 & 11.99 & 19.30
& 3.49 & 11.15 & 18.08 \\
FineLIP {\tiny [CVPR'25]}
& 3.72 & 13.05 & 20.51
& \underline{3.69} & 12.31 & 19.99 \\
SmartCLIP {\tiny [CVPR'25]}
& 2.13 & 8.65 & 14.39
& 2.08 & 7.93 & 13.45 \\
GOAL {\tiny [CVPR'25]}
& 3.52 & \underline{13.72} & \underline{21.72}
& 3.54 & \underline{13.23} & \underline{21.45} \\
StructXLIP {\tiny [CVPR'26]}
& \underline{3.81} & 12.48 & 20.71
& 3.27 & 12.43 & 20.36 \\
\midrule
\textbf{\ourmethod (Ours)}
& \textbf{4.33} & \textbf{14.14} & \textbf{22.90}
& \textbf{4.11} & \textbf{14.05} & \textbf{22.19} \\
\midrule
\textit{$\Delta$}
& \gain{0.52} & \gain{0.42} & \gain{1.18}
& \gain{0.42} & \gain{0.82} & \gain{0.74} \\
\bottomrule
\end{tabular*}
\end{table}

\begin{figure}[htbp] 
  \centering
  \includegraphics[width=\linewidth]{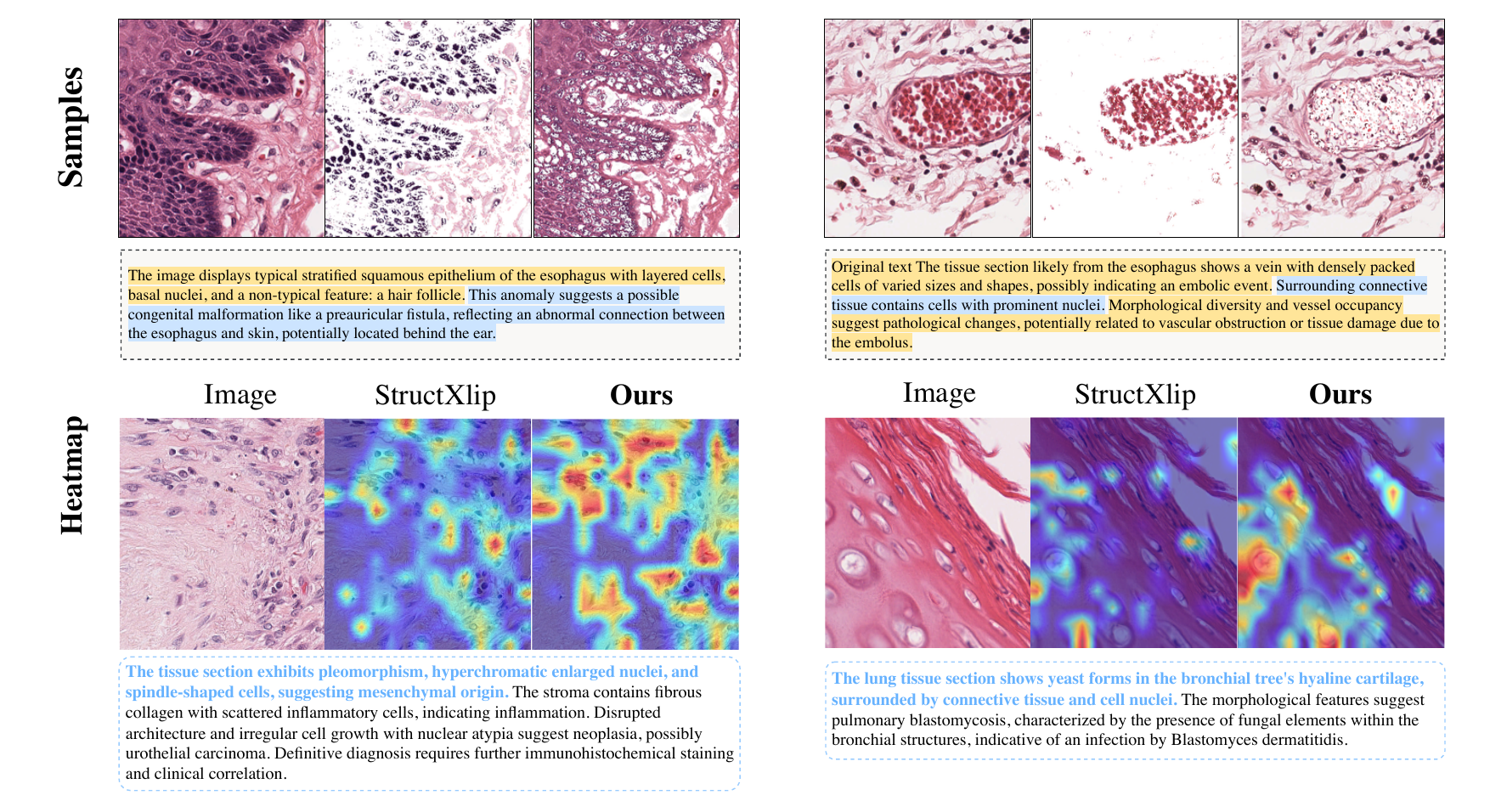} 
  \caption{
Qualitative examples from the PathGen benchmark after semantic condensation.
Top: representative pathology image-caption pairs and corresponding retrieval results.
Bottom: attention map comparisons between StructXLIP and our \ourmethod.
Our method \ourmethod exhibits better alignment with clinically relevant pathological regions described in the captions, including spindle-shaped tumor cells and yeast-like fungal structures associated with pulmonary blastomycosis.
}
  \label{fig:pathgen}
\end{figure}

Tab.~\ref{tab:pathgen} reports the cross-modal retrieval performance on PathGen. 
As a highly domain-specific benchmark with complex pathological patterns, PathGen remains challenging for all compared methods, resulting in relatively low absolute retrieval performance. Nevertheless, our method consistently achieves the strongest results among CLIP-based fine-tuning baselines, demonstrating the effectiveness of saliency-aware alignment in challenging pathology-specific scenarios.

Fig.~\ref{fig:pathgen} further provides qualitative comparisons on representative pathology samples.
Compared with StructXLIP, our method produces attention maps that better align with clinically relevant pathological regions described in the captions.
In the bottom-left example, the caption describes pleomorphic spindle-shaped tumor cells, a characteristic morphological pattern commonly associated with mesenchymal-origin malignancies. 
While StructXLIP mainly attends to scattered local regions without clearly capturing the spindle-cell morphology, our method accurately highlights the elongated spindle-shaped cellular structures mentioned in the report.
In the bottom-right example, the caption describes yeast forms consistent with pulmonary blastomycosis.
Our method successfully localizes most of the yeast-like fungal cells distributed within the bronchial cartilage region, whereas StructXLIP only focuses on several isolated regions and fails to capture the overall fungal distribution pattern.
Although minor background activations remain, our attention maps exhibit substantially better semantic correspondence with the pathological findings described in the caption.

\begin{table}[t]
\centering
\caption{
\textbf{Plug-and-play enhancement with our saliency-guided module on PathGen.}
We report Recall@K (\%) under both \inlineColorbox{lightpink}{\emph{Text$\rightarrow$Image}} and
\inlineColorbox{lightblue}{\emph{Image$\rightarrow$Text}} retrieval settings.
The values in parentheses denote the absolute change over the corresponding baseline.
}
\label{tab:pathgen-plug-in}
\scriptsize          
\setlength{\tabcolsep}{4pt}
\renewcommand{\arraystretch}{0.95}

\begin{tabular*}{\textwidth}{@{\extracolsep{\fill}} l *{2}{c!{\color{lightlightgray}\vrule width 0.5pt}}c|*{2}{c!{\color{lightlightgray}\vrule width 0.5pt}}c @{}}
\toprule
\multirow{2}{*}{\shortstack[c]{\textbf{Methods}}} &
\multicolumn{3}{c|}{\cellcolor{gray!15}\textbf{Text$\rightarrow$Image}} &
\multicolumn{3}{c}{\cellcolor{gray!15}\textbf{Image$\rightarrow$Text}} \\
\cmidrule(lr){2-4}\cmidrule(lr){5-7}
& \multicolumn{1}{c}{\cellcolor{lightpink}{R@1}}
& \multicolumn{1}{c}{\cellcolor{lightpink}{R@5}}
& \multicolumn{1}{c|}{\cellcolor{lightpink}{R@10}}
& \multicolumn{1}{c}{\cellcolor{lightblue}{R@1}}
& \multicolumn{1}{c}{\cellcolor{lightblue}{R@5}}
& \multicolumn{1}{c}{\cellcolor{lightblue}{R@10}} \\
\midrule
Long-CLIP
& 3.20 & 11.99 & 19.30
& 3.49 & 11.15 & 18.08 \\
\textbf{+our $\ourloss$}
& \textbf{3.67}\smallgain{0.47}
& \textbf{12.63}\smallgain{0.64}
& \textbf{20.44}\smallgain{1.14}
& 3.49{\scriptsize(+0.00)}
& \textbf{11.99}\smallgain{0.84}
& \textbf{19.74}\smallgain{1.66} \\
\midrule
FineLIP
& 3.72 & 13.05 & 20.51
& 3.69 & 12.31 & 19.99 \\
\textbf{+our $\ourloss$}
& \textbf{4.01}\smallgain{0.29}
& \textbf{13.15}\smallgain{0.10}
& 20.41{\smalldown{0.10}}
& \textbf{4.14}\smallgain{0.45}
& \textbf{13.10}\smallgain{0.79}
& \textbf{20.19}\smallgain{0.20} \\
\midrule
GOAL
& 3.52 & 13.72 & 21.72
& 3.54 & 13.23 & 21.45 \\
\textbf{+our $\ourloss$}
& \textbf{4.53}\smallgain{1.01}
& \textbf{15.61}\smallgain{1.89}
& \textbf{23.90}\smallgain{2.18}
& \textbf{4.21}\smallgain{0.67}
& \textbf{14.76}\smallgain{1.53}
& \textbf{23.28}\smallgain{1.83} \\
\midrule
DoRA
& 3.86 & 13.23 & 20.81
& 3.77 & 13.00 & 20.96 \\
\textbf{+our $\ourloss$}
& \textbf{4.24}\smallgain{0.38}
& \textbf{14.42}\smallgain{1.19}
& \textbf{21.92}\smallgain{1.11}
& \textbf{3.99}\smallgain{0.22}
& \textbf{14.37}\smallgain{1.37}
& \textbf{22.29}\smallgain{1.33} \\
\midrule
StructXLIP
& 3.81 & 12.48 & 20.71
& 3.27 & 12.43 & 20.36 \\
\textbf{+our $\ourloss$}
& \textbf{3.99}\smallgain{0.18}
& \textbf{13.95}\smallgain{1.47}
& \textbf{21.55}\smallgain{0.84}
& \textbf{3.57}\smallgain{0.30}
& \textbf{12.73}\smallgain{0.30}
& \textbf{20.86}\smallgain{0.50} \\
\bottomrule
\end{tabular*}
\end{table}

Tab.~\ref{tab:pathgen-plug-in} further demonstrates the plug-and-play capability of our module on PathGen. 
Our method consistently improves both CLIP-based fine-tuning approaches and PEFT-based adaptation methods, indicating strong compatibility and general applicability across different adaptation strategies.

\rebuttal{
\subsection{Fine-grained grounding}
\label{app:fine_grained_grounding}
}
We further evaluate fine-grained grounding
through three additional experiments: zero-shot pixel-level
localization, local-description retrieval, and local fine-grained discrimination.
None of these evaluations relies on our saliency-deletion protocol.

\textbf{Zero-shot pixel-level localization}
We evaluate referring image segmentation on RefCOCOg~\citep{mao2016generation}, where each sample consists of an image, a referring expression, and the ground-truth mask of the referred object. We use DCI-trained encoders and perform zero-shot evaluation without RefCOCOg or segmentation-specific training. Given a referring expression, we obtain the predicted mask using the same localization procedure for all compared models.
We report overall Intersection-over-Union (oIoU) and mean Intersection-over-Union (mIoU). oIoU computes the intersection and union over all evaluated samples globally, while mIoU averages the IoU of individual samples.
As shown in Tab.~\ref{tab:refcocog_grounding},
\ourmethod improves over StructXLIP by +0.91\% oIoU and +1.08\% mIoU, suggesting better zero-shot transfer to pixel-level localization on an unseen grounding benchmark.

\textbf{Fine-grained local-description retrieval}
We evaluate on DOCCI-FG from DOCCI dataset~\citep{docci2024}, which contains 5000 images and 35533 fine-grained local descriptions. Each local description is used to retrieve its corresponding full image from the complete gallery.
Tab.~\ref{tab:docci_fg_retrieval} shows 
\ourmethod consistently outperforms strong local-(GOAL) and structural-alignment(StructXLIP) baselines, showing improved use of localized fine-grained descriptions for image retrieval.


\begin{table}[t]
\centering

\begin{minipage}[t]{0.48\textwidth}
\vspace{0pt}
\centering

\caption{
\textbf{Zero-shot localization on RefCOCOg.}
We report oIoU and mIoU (\%).
We highlight \textbf{best} and \underline{second best} results;
$\Delta$ is our margin over StructXLIP.
}
\label{tab:refcocog_grounding}

\vspace{2pt}

\scriptsize
\setlength{\tabcolsep}{4pt}
\renewcommand{\arraystretch}{1.08}

\begin{tabular}{@{} l c c @{}}
\toprule

\textbf{Method}
& \textbf{oIoU}
& \textbf{mIoU} \\

\midrule

StructXLIP {\tiny [CVPR'26]}
& \underline{36.71}
& \underline{47.15} \\

\textbf{\ourmethod{} (Ours)}
& \textbf{37.62}
& \textbf{48.23} \\

\midrule

\textit{$\Delta$}
& \gain{0.91}
& \gain{1.08} \\

\bottomrule
\end{tabular}

\end{minipage}
\hfill
\begin{minipage}[t]{0.48\textwidth}
\vspace{0pt}
\centering

\caption{
\textbf{Local-description retrieval on DOCCI-FG.}
We report
\inlineColorbox{lightpink}{\emph{T$\rightarrow$I}}
Recall@K (\%).
We highlight \textbf{best} and \underline{second best} results;
$\Delta$ is our margin over the best competitor.
}
\label{tab:docci_fg_retrieval}

\vspace{2pt}

\scriptsize
\setlength{\tabcolsep}{4pt}
\renewcommand{\arraystretch}{1.08}

\begin{tabular}{@{} l c c c @{}}
\toprule

\textbf{Method}
& \textbf{R@1}
& \textbf{R@5}
& \textbf{R@10} \\

\midrule

GOAL {\tiny [CVPR'25]}
& \underline{18.01}
& \underline{34.19}
& \underline{41.81} \\

StructXLIP {\tiny [CVPR'26]}
& 16.61
& 31.47
& 38.81 \\

\textbf{\ourmethod{} (Ours)}
& \textbf{19.86}
& \textbf{37.46}
& \textbf{45.74} \\

\midrule

\textit{$\Delta$}
& \gain{1.85}
& \gain{3.27}
& \gain{3.93} \\

\bottomrule
\end{tabular}

\end{minipage}

\vspace{-5pt}
\end{table}

\textbf{Local fine-grained discrimination}
We evaluate on DCI's official local subset of 1,274
mask-aligned subcrops.
The models receive local subcrops without the full images
and must distinguish the correct description from negatives
differing in local attributes, parts, or relations.
Adding \ourmethod to StructXLIP improves accuracy by
3.77\%, 2.91\%, and 3.61\% 
under the standard, Pick5, and CLIP-hard-negative settings,
respectively.
These improvements support stronger discrimination of
fine-grained semantic differences using local visual content.



\section{Qualitative results}
\label{sec:qualitative}

\begin{figure}[htbp] 
  \centering
  \includegraphics[width=\linewidth]{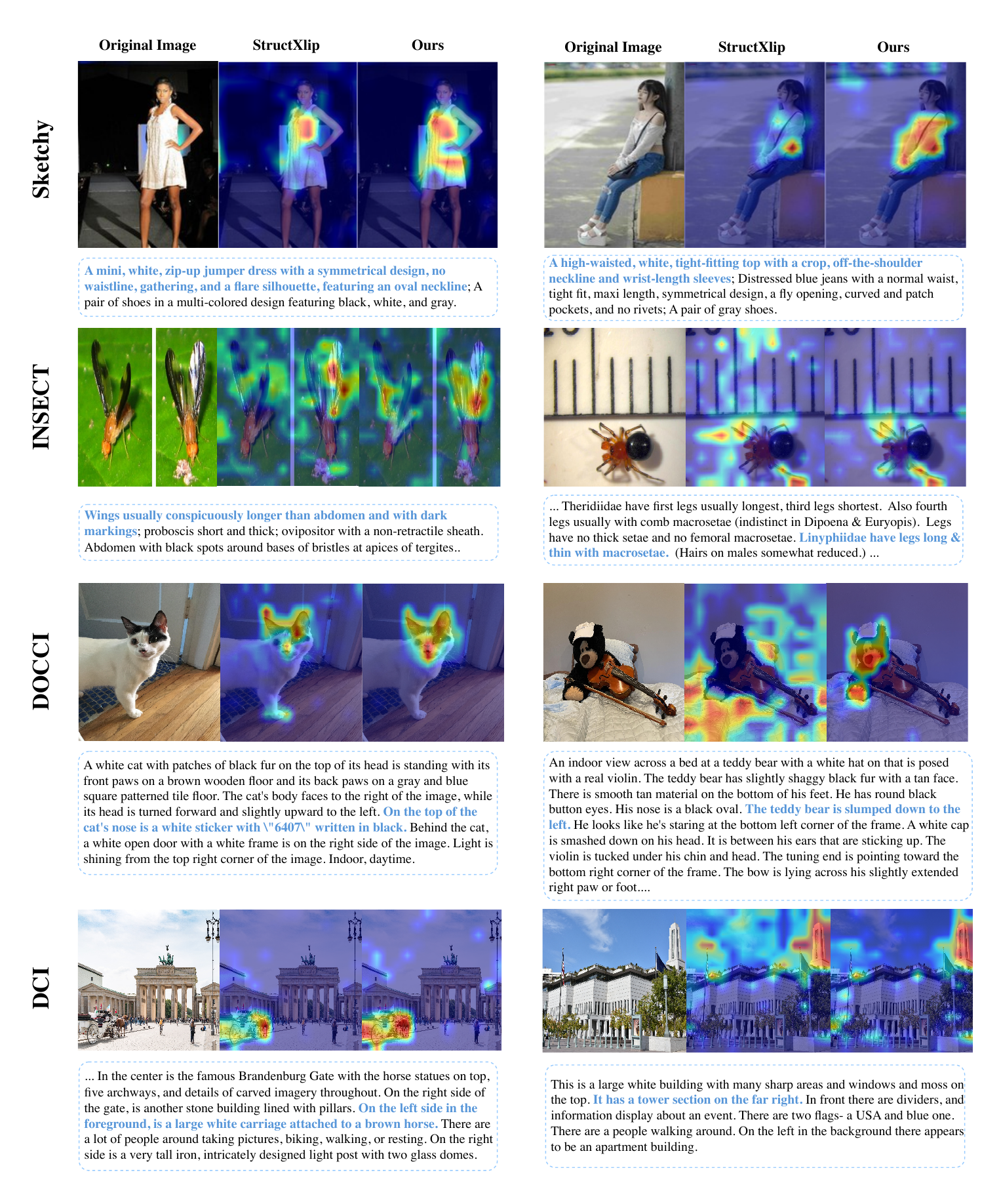} 
  
  \caption{\textbf{Qualitative results of \ourmethod} and the competitive method StructXLIP across four datasets. We visualize the attention maps corresponding to the \textcolor[HTML]{5796D9}{\textbf{blue bold}} text in each caption. Compared with StructXLIP, \ourmethod shows more focused activations on the corresponding visual objects described by the text.}
  \label{fig:attention}
\end{figure}

As shown in Fig.~\ref{fig:attention}, we further present qualitative comparisons between \ourmethod and the competitive StructXLIP~\citep{ruan2026StructXLIP} on four datasets: Sketchy~\citep{girella2025lots}, INSECT~\citep{truong2025insect}, DOCCI~\citep{docci2024} and DCI~\citep{urbanek2024picture}. Specifically, we visualize the attention maps on the image corresponding to the blue bold text in long captions. Overall, compared with StructXLIP, \ourmethod produces more focused and semantically consistent attention responses across different datasets, demonstrating stronger alignment between textual description and image regions.

On Sketchy dataset, both StructXLIP and \ourmethod can pay attention to the clothing, retaining a basic image-text alignment capability. However, the attention responses of StructXLIP are relatively localized and less stable. In contrast, \ourmethod provides more complete coverage of the target garments and better captures the overall clothing structures described in the text.
On INSECT dataset, since insect images belong to a relatively underrepresented visual domain, their categories and morphological structures are typically less sufficiently covered by the pre-training data of general vision-language models. As a result, the attention maps of both StructXLIP and \ourmethod are generally less accurate. Nevertheless, \ourmethod still presents better text-image alignment. 
On the DOCCI dataset, in the first case, the blue bold text explicitly describes a white sticker on the cat's nose with \'6407' written in black. Although StructXLIP roughly attends to the cat's head, its attention is mainly concentrated around the ears and fails to precisely localize the nose sticker described by the text. By contrast, \ourmethod places strong attention on the cat's nose region, providing a more accurate correspondence to the fine-grained text.






\newpage
\section*{NeurIPS Paper Checklist}

\begin{enumerate}

\item {\bf Claims}
    \item[] Question: Do the main claims made in the abstract and introduction accurately reflect the paper's contributions and scope?
    \item[] Answer: \answerYes{} 
    \item[] Justification: See section~\ref{sec:introduction}

\item {\bf Limitations}
    \item[] Question: Does the paper discuss the limitations of the work performed by the authors?
    \item[] Answer: \answerYes{} 
    \item[] Justification: See Section~\ref{sec:conclusion}

\item {\bf Theory assumptions and proofs}
    \item[] Question: For each theoretical result, does the paper provide the full set of assumptions and a complete (and correct) proof?
    \item[] Answer: \answerNA{} 
    \item[] Justification: The paper is empirical in nature and does not include theoretical results, assumptions, or formal proofs.

    \item {\bf Experimental result reproducibility}
    \item[] Question: Does the paper fully disclose all the information needed to reproduce the main experimental results of the paper to the extent that it affects the main claims and/or conclusions of the paper (regardless of whether the code and data are provided or not)?
    \item[] Answer: \answerYes{} 
    \item[] Justification: See Section~\ref{sec:experiments}

\item {\bf Open access to data and code}
    \item[] Question: Does the paper provide open access to the data and code, with sufficient instructions to faithfully reproduce the main experimental results, as described in supplemental material?
    \item[] Answer: \answerYes{} 
    \item[] Justification: Code and models will be made publicly available upon acceptance.

\item {\bf Experimental setting/details}
    \item[] Question: Does the paper specify all the training and test details (e.g., data splits, hyperparameters, how they were chosen, type of optimizer) necessary to understand the results?
    \item[] Answer: \answerYes{} 
    \item[] Justification: See Section~\ref{sec:experiments}

\item {\bf Experiment statistical significance}
    \item[] Question: Does the paper report error bars suitably and correctly defined or other appropriate information about the statistical significance of the experiments?
    \item[] Answer: \answerNA{} 
    \item[] Justification: The paper does not include error bars or statistical significance tests. All methods are evaluated under the same fixed benchmark splits and identical retrieval metrics, and the improvements are consistently observed across multiple datasets, backbones, and retrieval directions.

\item {\bf Experiments compute resources}
    \item[] Question: For each experiment, does the paper provide sufficient information on the computer resources (type of compute workers, memory, time of execution) needed to reproduce the experiments?
    \item[] Answer: \answerYes{} 
    \item[] Justification: See Section~\ref{sec:experiments}. We provide the compute worker types and memory used in our experiments, including a NVIDIA RTX 5090 GPU with 32GB memory and a NVIDIA A100 GPU with 40GB memory. We also report that each training run takes approximately 1–3 hours, depending on the dataset, and that results are averaged over three random seeds. The reported computation corresponds to the experiments included in the paper.
    
\item {\bf Code of ethics}
    \item[] Question: Does the research conducted in the paper conform, in every respect, with the NeurIPS Code of Ethics \url{https://neurips.cc/public/EthicsGuidelines}?
    \item[] Answer: \answerYes{} 
    \item[] Justification: The research fully complies with the NeurIPS Code of Ethics. It uses only public datasets and doesn't involve sensitive content.

\item {\bf Broader impacts}
    \item[] Question: Does the paper discuss both potential positive societal impacts and negative societal impacts of the work performed?
    \item[] Answer: \answerNA{} 
    \item[] Justification: The paper doesn't have any foreseeable societal impact. It aims to improve vision-language methods without involving human subjects, sensitive data, or deployment.
    
\item {\bf Safeguards}
    \item[] Question: Does the paper describe safeguards that have been put in place for responsible release of data or models that have a high risk for misuse (e.g., pre-trained language models, image generators, or scraped datasets)?
    \item[] Answer: \answerNA{} 
    \item[] Justification: The paper doesn't involve the release of data or models that have a high high risk for misuse. It is based on public available datasets.

\item {\bf Licenses for existing assets}
    \item[] Question: Are the creators or original owners of assets (e.g., code, data, models), used in the paper, properly credited and are the license and terms of use explicitly mentioned and properly respected?
    \item[] Answer: \answerYes{} 
    \item[] Justification: We use publicly available datasets and cite their original sources and licenses where applicable. We also rely on the official OpenCLIP codebase, which is released under the MIT License. All external resources are properly acknowledged and used in accordance with their respective licenses.

\item {\bf New assets}
    \item[] Question: Are new assets introduced in the paper well documented and is the documentation provided alongside the assets?
    \item[] Answer: \answerYes{} 
    \item[] Justification: We will release the model weights and codes upon acceptance with detailed usage information.

\item {\bf Crowdsourcing and research with human subjects}
    \item[] Question: For crowdsourcing experiments and research with human subjects, does the paper include the full text of instructions given to participants and screenshots, if applicable, as well as details about compensation (if any)? 
    \item[] Answer: \answerNA{} 
    \item[] Justification: The paper does not involve crowdsourcing or research with human subjects.

\item {\bf Institutional review board (IRB) approvals or equivalent for research with human subjects}
    \item[] Question: Does the paper describe potential risks incurred by study participants, whether such risks were disclosed to the subjects, and whether Institutional Review Board (IRB) approvals (or an equivalent approval/review based on the requirements of your country or institution) were obtained?
    \item[] Answer: \answerNA{} 
    \item[] Justification: The paper does not involve crowdsourcing or research with human subjects, and thus IRB or equivalent approval is not applicable.

\item {\bf Declaration of LLM usage}
    \item[] Question: Does the paper describe the usage of LLMs if it is an important, original, or non-standard component of the core methods in this research? Note that if the LLM is used only for writing, editing, or formatting purposes and does \emph{not} impact the core methodology, scientific rigor, or originality of the research, declaration is not required.
    \item[] Answer: \answerYes{} 
    \item[] Justification: See \supp Section~\ref{sec:generation_validation}

\end{enumerate}

\end{document}